\documentclass[preprint,sort&compress,12pt]{elsarticle}

\usepackage{amssymb}
\usepackage{amsthm}
\usepackage{amsmath}
\usepackage{mathtools}
\usepackage{mathrsfs}
\usepackage{algorithm}
\usepackage[algo2e]{algorithm2e}
\usepackage{algpseudocode}
\usepackage{array}
\usepackage{multirow}
\usepackage{listings}
\usepackage{tabu}
\usepackage{enumerate}
\usepackage{lineno}
\usepackage{fullpage}
\usepackage{xcolor}
\usepackage[colorinlistoftodos]{todonotes}
\usepackage{bbm}
\usepackage[colorlinks=true]{hyperref}
\usepackage{url}
\usepackage{textcomp}
\usepackage{gensymb}
\usepackage{soul}
\usepackage{graphicx}
\usepackage{subfig}
\usepackage{float}
\usepackage{caption}
\usepackage{tikz}
\usetikzlibrary{shapes}

\usepackage{wrapfig}
\usepackage{enumitem}

\theoremstyle{definition}

\theoremstyle{remark}

\biboptions{numbers,comma,round,square}
\graphicspath{ {./figs/} }

\journal{Elsevier}
\begin{document}
\begin{frontmatter}

 \title{Probabilistic Physics-integrated Neural Differentiable Modeling for Isothermal Chemical Vapor Infiltration Process}



\author[ndAME]{Deepak Akhare}
\author[ndCBE]{Zeping Chen}
\author[Honeywell]{Richard Gulotty}
\author[ndAME,Energy,ndCBE]{Tengfei Luo\corref{corjw}}
\author[ndAME,Energy,Lucy]{Jian-Xun Wang\corref{corjw}}


\address[ndAME]{Department of Aerospace and Mechanical Engineering, University of Notre Dame, Notre Dame, IN, USA}
\address[Energy]{Center for Sustainable Energy (ND Energy), University of Notre Dame, Notre Dame, IN, USA}
\address[ndCBE]{Department of Chemical and Biomolecular Engineering, University of Notre Dame, Notre Dame, IN, USA}
\address[Lucy]{Lucy Family Institute for Data \& Society, University of Notre Dame, Notre Dame, IN, USA}
\address[Honeywell]{Honeywell International Inc., Phoenix, AZ, USA}
\cortext[corjw]{Corresponding author: jwang33@nd.edu, tluo@nd.edu}

\begin{abstract}
Chemical vapor infiltration (CVI) is a widely adopted manufacturing technique used in producing carbon-carbon and carbon-silicon carbide composites. These materials are especially valued in the aerospace and automotive industries for their robust strength and lightweight characteristics. The densification process during CVI critically influences the final performance, quality, and consistency of these composite materials. Experimentally optimizing the CVI processes is challenging due to long experimental time and large optimization space. To address these challenges, this work takes a modeling-centric approach. Due to the complexities and limited experimental data of the isothermal CVI densification process, we have developed a data-driven predictive model using the physics-integrated neural differentiable (PiNDiff) modeling framework. An uncertainty quantification feature has been embedded within the PiNDiff method, bolstering the model's reliability and robustness. Through comprehensive numerical experiments involving both synthetic and real-world manufacturing data, the proposed method showcases its capability in modeling densification during the CVI process. This research highlights the potential of the PiNDiff framework as an instrumental tool for advancing our understanding, simulation, and optimization of the CVI manufacturing process, particularly when faced with sparse data and an incomplete description of the underlying physics.
\end{abstract}

\end{frontmatter}


\section{Introduction}
\label{sec:intro}

%

Carbon-carbon (C/C) and carbon-silicon carbide (C/SiC) composites are recognized as vital materials for applications exposed to extreme thermal conditions, notably in contexts such as airplane brake discs, reentry vehicle heat shields, and rocket engine nozzles. These composites are distinguished by their exceptional thermal stability and superior mechanical properties under high temperatures. Their fabrication is largely achieved via the isothermal chemical vapor infiltration (I-CVI) process, which is known for its precision in defining final properties, its adaptability in accommodating complex geometries, and its ability to yield final products that are both lightweight and thermally robust. Since the performance, consistency, and quality of these composites are largely influenced by the manufacturing procedures, optimizing the I-CVI process is of great importance~\cite{delhaes2002chemical,fu2022micro}. However, this is not a trivial task and presents significant challenges, given the long duration of I-CVI, often lasting months, rendering conventional trial-and-error approaches infeasible~\cite{zhao2022high}.

The inherently time-consuming nature of the I-CVI process necessitates the creation of a predictive model that enables efficient computer-based simulations. This strategic shift offers the potential for markedly reduced turnover times, allowing for a comprehensive exploration of varied manufacturing conditions. A reliable and efficient computer-based predictive model can enable us to effectively optimize the I-CVI process, exerting finer control over its intricacies, with the aim of reducing production cycles and ensuring resultant materials meet targeted properties. Although several numerical models, particularly those focusing on Carbon and SiC deposits, have been developed for the I-CVI process in the literature~\cite{vignoles2015modeling, kim2021full,wei2006two,wei2006numerical1, wei2006numerical2}, they face inherent challenges. These models often rely on multiple assumptions due to unresolved/unknown physical phenomena, which compromise their accuracy and reliability. Most of the time, these models are tailored to specific processes, limiting their adaptability to other new process designs. Given the multifaceted physical interactions of I-CVI and its nuanced mechanical and chemical mechanisms, a purely physics-based modeling approach appears impractical and non-scalable. Moreover, the development and simulation of these models demand substantial computational resources, restricting their immediate applicability for optimization and uncertainty assessments.

To effectively simulate and optimize the I-CVI process, the creation of a new modeling framework that is both generally applicable and scalable is indispensable. The growing availability of data offers promising avenues for machine learning (ML)-based, data-driven modeling techniques to make substantial advancements in this regard. Deep neural networks (DNNs), cornerstones of ML, have long been recognized for their efficacy in diverse fields, ranging from speech recognition and image analysis to natural language processing. Harnessing the rapid advancements in computational resources and algorithmic developments, along with a vast accumulation of experimental and simulation data, DNNs have been established as essential tools in the domain of scientific modeling and simulation. Their impact is notably evident in the field of scientific machine learning (SciML), which is progressively being adopted in predictive composite manufacturing~\cite{WANG202213,LIU2021109152,HUANG2021113917}. Through SciML, researchers have been able to identify previously unknown constitutive relationships of composite materials and have optimized the speed and efficiency of multiscale simulations~\cite{LIU2021109152}. Wang et al.~\cite{WANG202213}, for instance, utilized ML techniques to analyze patterns within additive manufacturing datasets and developed models that describe process-structure-property relationships across different parameters. Huang et al.~\cite{HUANG2021113917} highlighted the advantages of their data-driven model, which was trained on a previously published experimental dataset, in predicting the mechanical characteristics of carbon nanotube-reinforced cement composites, especially when compared to traditional response surface methods. Nguyen et al.~\cite{nguyen2022use} investigated the impact of curing on stress development and tensile transverse failure response by integrating a thermo-chemo-mechanical finite element model with a DNN-based constitutive model. This allowed them to understand matrix mechanical property changes in relation to temperature and curing levels. Kopal et al.\cite{kopal2022generalized} employed regression neural networks to predict the curing properties of rubber blends enriched with carbon black, considering both blend types and curing temperatures. Baek et al.\cite{baek2022deep} employed graph neural networks to study the effects of nanoparticle distribution patterns and agglomeration phenomena in polymers. Li et al.\cite{li2003modeling} designed a data-driven DNN model to learn the nonlinear interdependencies between I-CVI process parameters and the resulting physical properties of C/C composites, which can be used for process optimization.

While these pure data-driven DNN methods have been successful across diverse applications in composite manufacturing modeling, there is a noticeable gap in the literature concerning their use for modeling the CVI process. One of the primary obstacles is the substantial data requirement intrinsic to these models. Collecting sufficient data to effectively train and utilize a purely data-driven DNN model for CVI optimization can be extremely challenging, if not infeasible. Another limitation is the restricted generalizability of these models, where failures were often observed beyond the training regimes~\cite{LIU2021109152}, rendering them less suited for complex CVI processes that operate under a spectrum of manufacturing conditions. The physics-informed deep learning (PIDL) strategy, leveraging physics principles to enhance the design, training, or inference of DNNs, presents a promising potential solution to these challenges. It is worth noting that the PIDL approach has garnered attention and demonstrated considerable potential across various domains, including solid mechanics~\cite{haghighat2021physics,gao2022physics,abueidda2021meshless}, turbulent flows~\cite{wang2017physics,yang2019predictive,duraisamy2019turbulence}, materials~\cite{zhang2022analyses}, heat transfer~\cite{cai2021physics,niaki2021physics,li2022physics,zobeiry2021physics,li2023physics,luo2023physics,li2021physics} 
, and biomechanics~\cite{kaandorp2021improved,arzani2022machine,sarabian2022physics}. A notable PIDL method is the Physics-informed neural network (PINN)~\cite{cuomo2022scientific,raissi2019physics,sun2020surrogate,laubscher2021simulation,henkes2022physics,niaki2021physics}, wherein the governing equations are incorporated into the loss function to regularize the training process, thereby reducing the required labeled data. However, the introduction of nonlinearity into the loss function can pose significant challenges in optimization~\cite{wang2022and}, and the physics-based loss function requires complete governing equations of the underlying physics, which are not available for the intricate I-CVI processes. 

To integrate incomplete physics with deep learning, the hybrid differentiable neural modeling emerges as a notable alternative PIDL method, which fuses physics-derived mathematical models with the robust learning capability of neural networks, ensuring efficient learning even with limited data. Differentiable programming (DP) serves as a cornerstone in this framework, allowing for the joint optimization of both DNNs and physics-centric components within a unified training environment. The development of differentiable physics solvers and hybrid neural models has recently gained traction, exemplifying their adaptability across various scientific fields~\cite{innes2019differentiable,belbute2020combining,huang2020learning,kochkov2021machine,list2022learned,akhare2023physics,liu2022predicting,fan2023differentiable}. Notably, Akhare et al.~\cite{akhare2023physics} introduced a {physics-integrated neural differentiable (PiNDiff) framework}, developed for the curing process of composites. This approach seamlessly integrates partially-known physics into neural networks while preserving the mathematical integrity of governing equations through DP. Akhare and co-workers demonstrated the effectiveness of the PiNDiff method in capturing the interplay between heat transfer and curing dynamics~\cite{akhare2023physics}. Driven by its potential, our objective in this work is to employ the PiNDiff framework to develop a predictive model for the I-CVI process in fabricating C/C composites, highlighting its ability to work with partially-known physics and sparse indirect measurements. Extending from its original formulation, our study broadens the PiNDiff's scope to the I-CVI process, governed by both hyperbolic and elliptic partial differential equations (PDEs). A primary thrust of this effort is to distill the existing physics-based model into a more concise PDE system that serves as the physics-based backbone of the PiNDiff model. However, this simplification, along with the potential over-parameterization of neural operator components, may introduce errors in predictions. To address this issue, we extend the PiNDiff framework by integrating uncertainty quantification (UQ) features, utilizing Deep Ensemble (DeepEn) techniques~\cite{lakshminarayanan2017simple}. Consequently, our probabilistic PiNDiff I-CVI model is not only able to learn and predict I-CVI processes with limited training data but also to gauge the confidence of its predictions. This development facilitates a more comprehensive evaluation of the model's reliability, thereby guiding decision-making processes more adeptly. The rest of the paper is organized as follows: the overall methodology of the PiNDiff I-CVI model is introduced in Section~\ref{sec:methodology}. Numerical experiments of the PiNDiff I-CVI model and its comparison with the experimental data are presented in Section~\ref{sec:result}. Finally, Section~\ref{sec:conclusion} concludes the paper.

\section{Methodology}
\label{sec:methodology}

\subsection{Isothermal Chemical Vapor Infiltration (I-CVI) process: an overview}

CVI stands as a pivotal process in the production of composite materials, notably C/C and C/SiC composites. The core principle behind this process is the infiltration of a porous preform using reactive gases, which undergo chemical reactions to deposit solid material into the interstitial spaces of the preform's structure (known as the matrix), thereby resulting in a reinforced composite. The isothermal CVI variant, known as I-CVI, distinguishes itself by maintaining a constant temperature throughout the infiltration process.
A typical I-CVI procedure unfolds as follows:
\begin{itemize}
    \item Preform Design and Creation: Using materials such as carbon fibers, a porous preform is meticulously crafted. Its structure is designed to embody the desired shape, serving as the foundational framework for the final composite.
    \item Reactor Chamber Setup: The crafted preform is situated within a specialized reactor chamber for the I-CVI procedure. This chamber is not only constructed to endure high temperatures but also engineered to regulate gas flow during infiltration.
    \item Gas Selection and Introduction: Depending on the intended composition of the resulting composite, specific reactive gases, such as hydrocarbons or silanes, are channeled into the chamber. These gases contain the essential elements required for the desired chemical reactions and material deposition.
    \item Chemical Reactions \& Material Deposition: As these gases permeate the preform, they undergo thermal reactions, especially on the surface of the preform. Such chemical reactions facilitate the progressive deposition of solid material--commonly carbon or a carbon-silicon carbide mix--into the matrix of the preform layer by layer. This deposition methodically fills the preform's voids, thereby reinforcing its structure.
    \item Regulation of Temperature and Duration: The I-CVI process is generally executed at a consistently elevated temperature, frequently between $900^o$ C to $1200^o$ C. Both the temperature and the process duration are meticulously regulated to guarantee accurate deposition and desired material characteristics. Depending on the intended composite thickness and properties, the procedure's duration can range from a few hours to multiple days.
    \item Post-Process Refinement/Machining: Following the main I-CVI procedure, the produced composite might necessitate further refinements, such as surface treatments or precise machining, to fine-tune its properties and dimensions.
\end{itemize}

\begin{figure}[!ht]
\centering
\includegraphics[width = 0.9\textwidth]{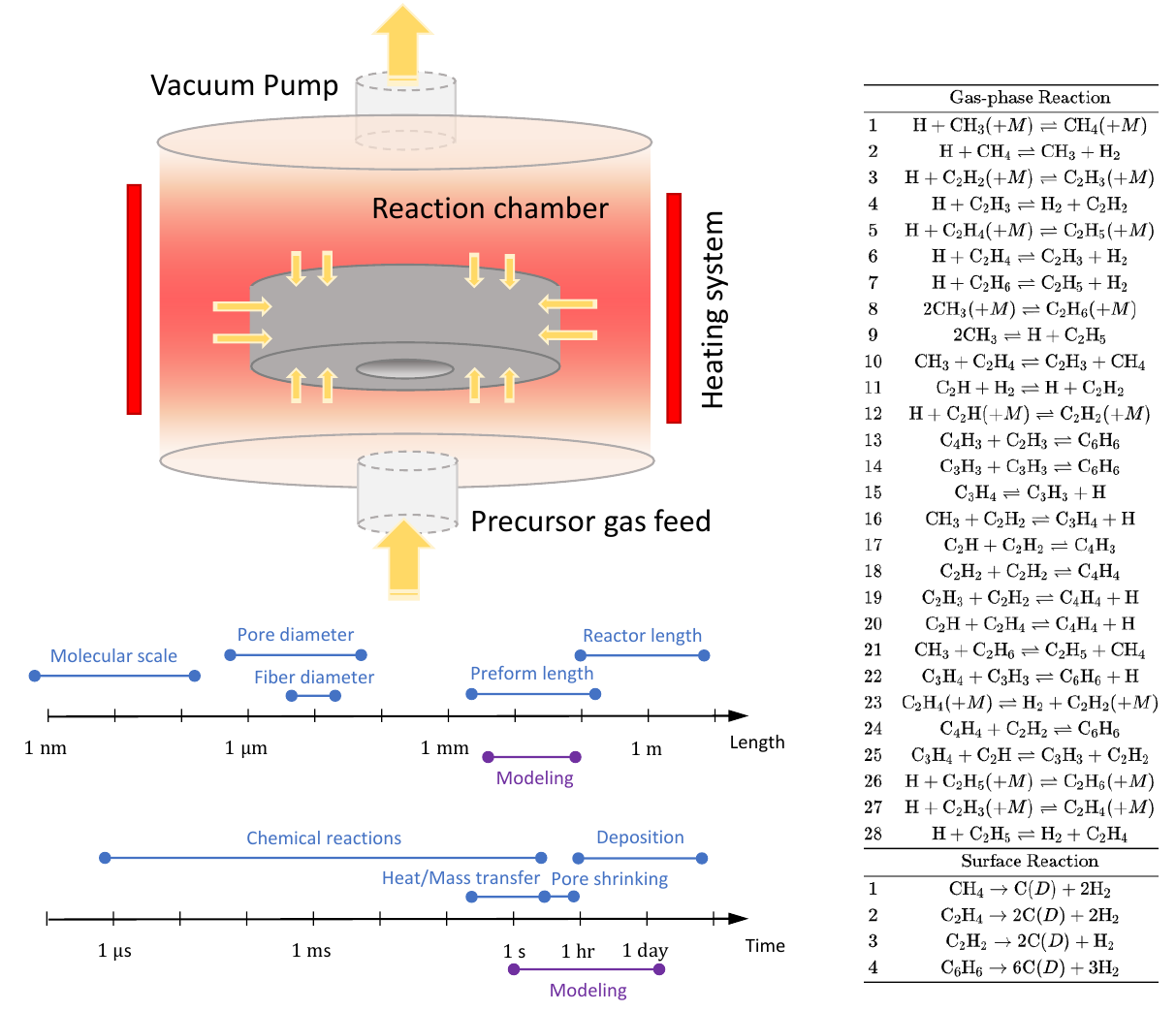}
\caption{A schematic illustrating the I-CVI process. Depicted are the dominant chemical reactions specific to hydrocarbon deposition, integrated with visual representations of the varying scales of multiphysical and multichemical phenomena inherent in the process.}
\label{fig:CVIprocess}
\end{figure}

The I-CVI process involves complex multiphysical and multichemical phenomena operating over multiple spatial and temporal scales. These include fluid dynamics, homogeneous gas-phase reactions, as well as intricate heterogeneous reactions leading to surface deposition. A comprehensive schematic illustrating the dominant chemical reactions for hydrocarbon deposition, along with the varying scales of these multiphysics phenomena, is presented in Fig.~\ref{fig:CVIprocess}.

\subsection{Physics-based modeling for I-CVI process and challenges}
\label{subsec:Num_model}
The I-CVI process encompasses multiple interconnected physical phenomena, including gas flow, diffusion, and both homogeneous and heterogeneous reactions, which can be mathematically modeled as a set of partial differential equations (PDEs) describing the mass and momentum conservation~\cite{blake1976species,mcallister1993simulation}:
\begin{itemize}
\begin{subequations}
\item Mass balance (reaction-diffusion): This equation portrays the interplay between the reaction and diffusion of gas-phase species,
  \begin{equation}
    \frac{\partial(\varepsilon_k\rho_k)}{\partial t} + \nabla \cdot (\varepsilon_k\rho_k\textbf{u}) = D_{k} \nabla^2(\varepsilon_k\rho_k) + \dot{\omega}_k \ \ k=1,...,N_g.
  \end{equation}
\item Mass balance (deposition): This equation captures the dynamics of species that contribute to deposition,
  \begin{equation}
    \frac{\partial(\varepsilon_k\rho_k)}{\partial t} + \nabla \cdot (\varepsilon_k\rho_k\textbf{u}_k) = D_{k} \nabla^2(\varepsilon_k\rho_k) + \dot{\omega}_k \ \ k=1,...,N_s.
  \end{equation}
\item Momentum balance for gas species: This encapsulates the conservation of momentum for gas-phase species:
  \begin{equation}
    \frac{\partial(\rho_g\textbf{u})}{\partial t} + \nabla \cdot (\rho_g\textbf{u}\textbf{u}) = \nabla.\boldsymbol{\sigma } + \rho_g\textbf{g} + \textbf{F}.
  \end{equation}
\end{subequations}
\end{itemize}
In the above equations, $N=N_g+N_s$ is the total number of species, where $N_g$ is the number of species in the gas phase and $N_s$ is the number of species resulting in deposition; $\varepsilon_k$ represents the volume fraction of the $k$-th species, $\rho_k$ and $\rho_g$ is the average intrinsic mass density of the $k$-th species and gas, $D_{k}$ is the diffusion coefficient of the $k$-th species, and $\dot{\omega}_k$ is the production rate of the $k$-th species; $\textbf{u}$ represents the gas velocity, $\boldsymbol{\sigma}$ denotes the total stress tensor, $\textbf{g}$ is the external body force per unit mass, and $\textbf{F}$ is the macro-scale momentum force. 

While the I-CVI process can be described by an extensive set of physics principles, using numerical techniques to simulate these intricate phenomena poses substantial challenges in practice. At the forefront of these challenges is the need for a comprehensive understanding of the dominant reactions and key parameters, such as the collision integral, crucial for determining the diffusion coefficient $D_{k}$ for all species. Such detailed information, however, often remains elusive in many real-world applications. Moreover, translating these foundational physics concepts into predictive simulations demands numerically solving N+1 sets of mass and momentum PDEs to capture the behavior of N species, thereby compounding the complexity. A noteworthy caveat is that the most rigorous physics-based models often fall short in capturing the actual measurements, possibly due to an incomplete understanding of the underlying physics. For instance, Kang et al.~\cite{guan2018numerical} compared the experimental results of SiC densification with predictions from physics-based numerical models that incorporated a variety of experimentally established reaction equations. The models, however, failed to accurately match the experimental data. On successfully constructing a numerical model, one is still confronted with its intractable computational costs, making it infeasible for tasks such as optimization and UQ. Additionally, these models are often tailored for a specific precursor gas, limiting their applicability to that particular substance and requiring reconfiguration for alternative precursor gases.

\subsection{Probabilistic physics-integrated neural differentiable modeling for I-CVI process}

To address the challenges highlighted above, the physics-integrated neural differentiable (PiNDiff) modeling framework, as proposed by Akhare et al.~\cite{akhare2023physics}, emerges as an innovative approach for constructing an effective predictive surrogate model for the I-CVI process. This strategy leverages both the established knowledge of physics and the insights learned from sparse, indirect data. Central to the PiNDiff modeling framework is its seamless fusion of deep learning capabilities with foundational physics principles, resulting in a hybrid neural solver. Within the PiNDiff module, trainable DNNs are utilized to learn unresolved/undetermined physics, while the non-trainable networks are pre-determined by the PDE operators from the partially known physics. Notably, the entire architecture of this hybrid neural solver is fully differentiable, facilitating its holistic training and optimization. To account for data scarcity and inherent model-form error, the PiNDiff framework is extended in this work to equip it with UQ capability. 

\subsubsection{PiNDiff model architecture design}
To construct a PiNDiff surrogate model for the I-CVI process, the major foundational physics can be distilled based on a set of assumptions to streamline the modeling process. Specifically, (a) the flow outside the fibrous preform is considered to have negligible influence on the diffusion process or the concentration of the precursor gas at the boundary of the preform; (b) flow within the fibrous preform is assumed to be negligible, suggesting that the transportation of the precursor gas is predominantly through diffusion; and (c) instead of simulating multiple species, an aggregate representation is used, introducing an effective species responsible for both diffusion and deposition processes~\cite{wei2006two}.
Operating under these assumptions allows for a more simplified neural modeling strategy. Specifically, the flow outside the preform becomes irrelevant to the model, restricting the domain of interest solely to the preform itself. Additionally, the foundational physics is substantially reduced to two PDEs: one for reaction-diffusion and another for deposition. By invoking a quasi-steady-state approximation, the time derivative term in the reaction-diffusion equation can be eliminated, further simplifying the governing equations~\cite{wei2006two}. For the deposition reactions, a first-order rate expression, given by $\dot{\omega}=-M_dKS_vC$, is adopted~\cite{wei2006two, guan2018numerical}. Finally, the two PDEs can be derived as follows (see derivation details in \ref{sec:app-derivation}),\\
\begin{subequations}
Reaction-Diffusion mass balance equation:
  \begin{equation}
    D_{eff} \nabla^2( C) = KS_vC,
  \end{equation}
Deposition mass balance equation:
  \begin{equation}
    \rho_d\frac{d\varepsilon}{dt} = - qM_d KS_vC.
  \end{equation}
  \label{eq:I-CVI-simp}
\end{subequations}
In the above equations, $C = C(\mathbf{x}, t)$ denotes the effective molarity field (mol m$^{-3}$) of all reactive gases, $\varepsilon = \varepsilon(\mathbf{x}, t)$ is the porosity of the preform, $q$ represent a constant stichometric coefficient, $M_d$ is the molar mass (kg mol$^{-1}$), and $\rho_d$ is the density (kg m$^{-3}$) of the deposited solid (carbon or SiC). $D_{eff} = D_{eff}(\mathbf{x}, t)$ represents the effective diffusion coefficient field, $K = K(\mathbf{x}, t)$ is the deposition reaction rate, and $S_v = S_v(\mathbf{x}, t)$ corresponds to the surface-to-volume ratio. 

\begin{figure}[ht]
\centering
\includegraphics[width = 0.8\textwidth]{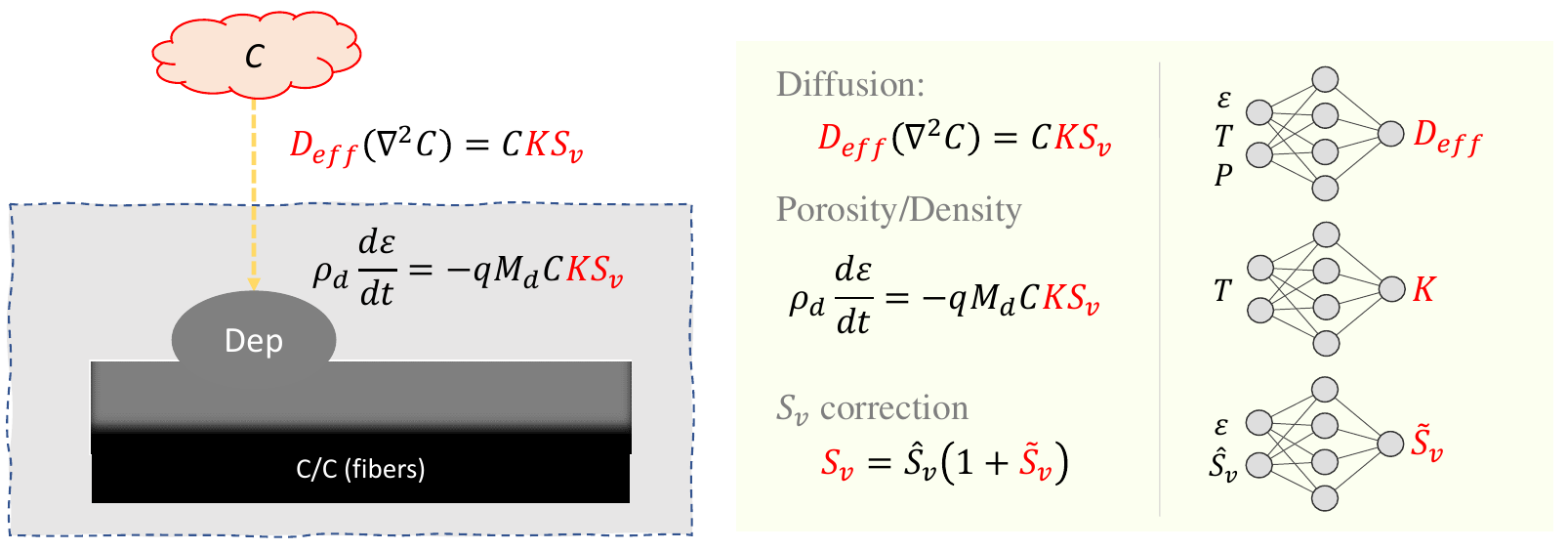}
\caption{Extracted foundational physics of the I-CVI process with neural operator approximations.}
\label{fig:CVI_model}
\end{figure}
Deriving constitutive relations for the spatio-temporal functions  $D_{eff}(\mathbf{x}, t)$, $K(\mathbf{x}, t)$, and $S_v(\mathbf{x}, t)$ poses significant challenging. Specifically, the models for $D_{eff}$ and $K$ are sensitive to the choice of precursor gas, while $S_v$ is influenced by the architecture of fibers and pores. Given these intricacies, we propose to learn these constitutive relations from sparse measurements within the PiNDiff framework. Namely, these constitutive relations are essentially unknown mappings between different spatio-temporal functions, which can be approximated by trainable neural operators. Specifically, the effective diffusion coefficient field $D_{eff}(\mathbf{x}, t)$ and deposition rate field $K(\mathbf{x}, t)$ are modeled as,
\begin{linenomath*}
\begin{subequations}
    \begin{alignat}{3}
    D_{eff}(\mathbf{x}, t) &\approx \mathcal{D}_{nn}\big[\varepsilon(\mathbf{x}, t), T(\mathbf{x}, t), P(\mathbf{x}, t); \boldsymbol{\theta}_{D_{eff}}\big],\\
    K(\mathbf{x}, t) &\approx \mathcal{K}_{nn}\big[T(\mathbf{x}, t); \boldsymbol{\theta}_{K}\big], 
    \end{alignat}
    \label{eq:D_K_approx}
\end{subequations}
\end{linenomath*}
where $\mathcal{D}_{nn}$ and $\mathcal{K}_{nn}$ are neural operators with trainable parameters $\boldsymbol{\theta}_{D_{eff}}$ and $\boldsymbol{\theta}_K$, respectively. 
As for the effective surface area $S_v(\mathbf{x}, t)$, an analytical model, $\hat{S}_v(\mathbf{x}, t)$, is leveraged as the base model. To accommodate a broader range of fiber configurations, a trainable neural operator $\mathcal{S}_{nn}$ with trainable parameters $\boldsymbol{\theta}_{S_v}$, is introduced to augment and generalize the base model as follows:  
\begin{subequations}
    \begin{alignat}{3}
    S_v(\mathbf{x}, t) &\approx \hat{S}_v(\mathbf{x}, t) (1 + \tilde{S}_v(\mathbf{x}, t)),\\
    \tilde{S}_v(\mathbf{x}, t) &= \mathcal{S}_{nn}[\hat{S}_v(\mathbf{x}, t), \varepsilon(\mathbf{x}, t); \boldsymbol{\theta}_{S_v}],\\
    \hat{S}_v(\mathbf{x}, t) &= 2 \frac{\varepsilon}{\varepsilon_0} \frac{1-\varepsilon_0}{r_f}
    \end{alignat}
    \label{eq:Sv_approx}
\end{subequations}
where $\varepsilon_0$ is the initial porosity of the preform and $r_f$ is the radius of fibre filament (m). 

\begin{figure}[ht]
\centering
\subfloat[Auto-regressive PiNDiff prediction model for I-CVI process model]{\includegraphics[width = 0.95\textwidth]{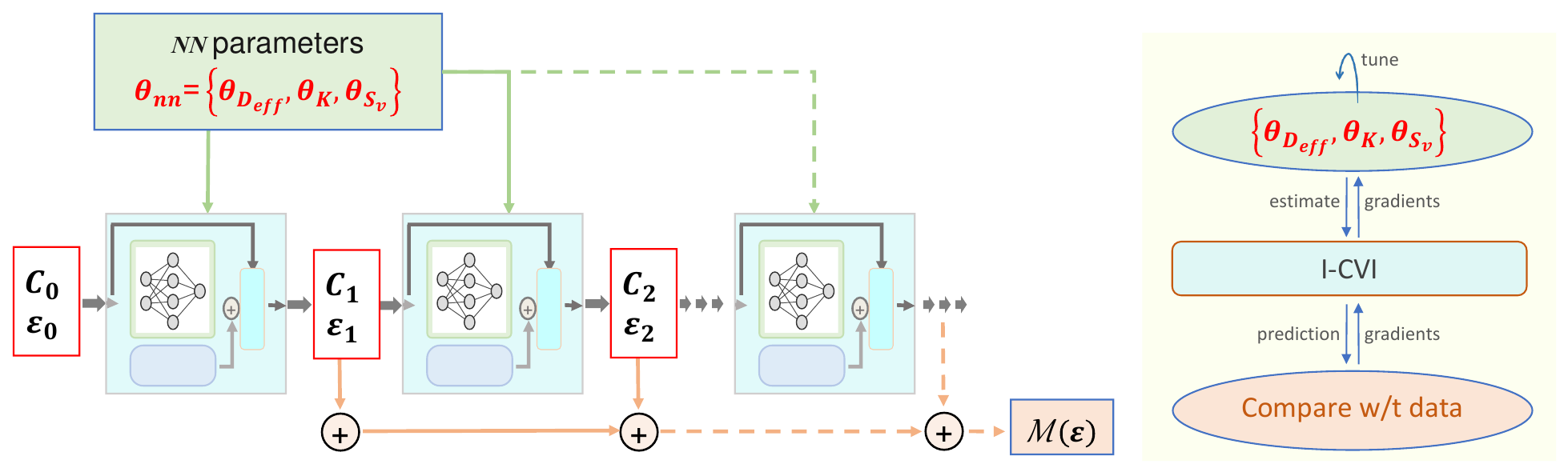}}\\
\subfloat[one-step PiNDiff module for I-CVI process model]{\includegraphics[width = 0.95\textwidth]{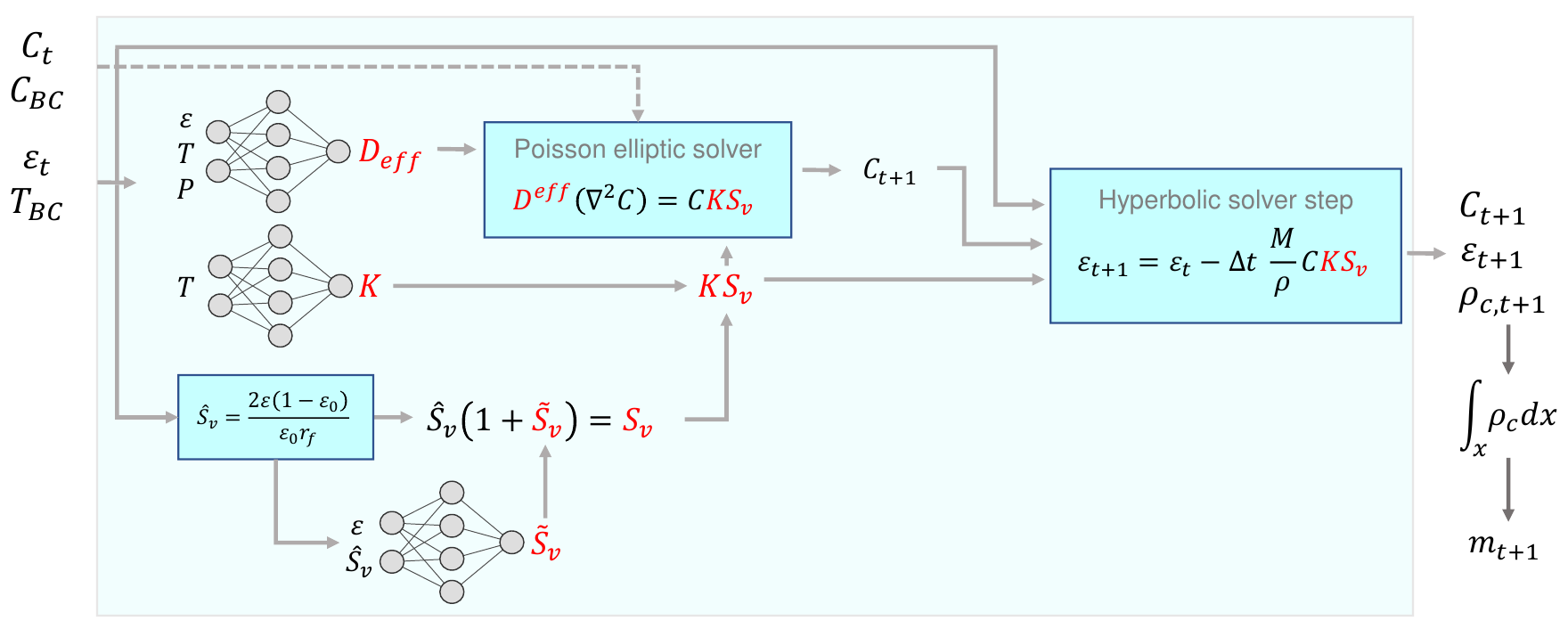}}
\caption{(a) The overview of the auto-regressive learning architecture of the PiNDiff model and (b) Zoom-in view of the PiNDiff module for one time-step prediction (fu: function).}
\label{fig:PiNDiffnet}
\end{figure}

In Fig.~\ref{fig:PiNDiffnet}, the PiNDiff I-CVI model is presented, revealing a seamless integration of known physics with trainable neural operators. These operators are strategically interconnected, drawing inspiration from the foundational principles delineated by the simplified governing PDEs in Eq.\ref{eq:I-CVI-simp}. Within this architectural framework, the DNNs operate in a pointwise fashion, effectively learning continuous functions/operators. This design ensures both mesh and domain independence, enhancing the model's adaptability and versatility. 
As the building blocks, the PiNDiff I-CVI module (Fig.~\ref{fig:PiNDiffnet}b) operates as a sequential neural predictor, capturing the evolution of molarity ($C$), porosity ($\varepsilon$), and composite density ($\rho_c$) fields from time step $t$ to $t+1$. This module incorporates trainable neural operators designed to approximate the spatiotemporal fields, namely $D_{eff}(\mathbf{x}, t)$, $K(\mathbf{x}, t)$, and $S_v(\mathbf{x}, t)$. These operators are interlinked by the discretized governing PDEs with finite difference methods. A unique aspect of this design is the harmonious merger of neural networks and numerical PDEs via differentiable programming, ensuring efficient gradient propagation throughout the model (detailed in the next subsection). Upon domain-wide integration, the model is capable of predicting pertinent variables, notably the mass or weight of the materials involved. The ability to predict such metrics is of significant importance, especially given that monitoring material mass/weight is a common practice in manufacturing processes. Leveraging the PiNDiff I-CVI module, we can construct sequential neural networks using auto-regressive techniques, adeptly capturing the system's temporal dynamics, as illustrated in Fig.~\ref{fig:PiNDiffnet}a, highlighting the structured sequential learning for the I-CVI process. 

\subsection{Auto-regressive training of PiNDiff model via differentiable programming}

To ensure its robust long-term forecast capabilities, the PiNDiff I-CVI model undergoes auto-regressive training throughout the entire rollout sequence, which is achieved by differentiable programming ($\partial$P)~\cite{innes2019differentiable}, a generalized concept of deep learning. Namely, the entire computer program is architected for end-to-end differentiability. In the construction of the PiNDiff I-CVI model, we utilize the automatic differentiation (AD) engine provided by JAX~\cite{jax2018github} to propagate gradients through the hybrid model, enabling the optimization of all trainable parameters, denoted as $\boldsymbol{\theta}= \{\boldsymbol{\theta}_{D_{eff}}, \boldsymbol{\theta}_K, \boldsymbol{\theta}_{S_v} \}$, through stochastic gradient descent techniques. Notably, in contrast to traditional deep learning paradigms, this differentiable framework offers the flexibility to train the model using indirect labels, even if the state variables of interest are not directly observable.

In the context of I-CVI, experimentally monitoring the states such as porosity $\varepsilon(\mathbf{x}, t)$ and molarity $C(\mathbf{x}, t)$ during the manufacturing process is impractical. Meanwhile, hidden physics such as effective diffusion, reaction, and deposition mechanisms are not observable at all. Nonetheless, acquiring measurements for the mass of either the entire sample or its segments is considerably more feasible. Consequently, mass measurements ($\tilde{\mathbf{m}}_t$) at a few different time step $t$ serves as indirect labeled data for training the PiNDiff I-CVI model. The associated loss function, $\mathcal{L}$, is then defined as,
$$ \mathcal{L}(\boldsymbol\theta) = \sum_{t=0}^{t_n} \Big\lVert \mathcal{M}\big({\textbf{v}_t(\bar{\textbf{x}}|\boldsymbol\theta)}\big)-\mathbf{\hat{m}}_t \Big\lVert_{L_2} + \beta_1 \big\lVert \boldsymbol{\theta}\big\lVert_{L_2} + \beta_2 \sum_{t=0}^{t_n-1} \Big\lVert\textbf{v}_{t+1}(\bar{\textbf{x}}|\boldsymbol\theta) -\textbf{v}_t(\bar{\textbf{x}}|\boldsymbol\theta)\Big\lVert_{L_2} $$
\begin{equation}
    + \beta_3 \Big\lVert D_{eff}(\bar{\textbf{x}}|\boldsymbol\theta)\Big(\nabla^2C(\bar{\textbf{x}}) - C(\bar{\textbf{x}})K(\bar{\textbf{x}}|\boldsymbol\theta)S_v(\bar{\textbf{x}}|\boldsymbol\theta)\Big)\Big\lVert_{L_2},
    \label{eq:loss}
\end{equation}
where $\mathcal{M}: \mathbf{v} \to \mathbf{m}$ is the state-to-observable map, which maps from the state variables $\mathbf{v} = [\varepsilon(\mathbf{x}, t), \rho_c(\mathbf{x}, t), C(\mathbf{x}, t) ]$ to the observables (e.g, mass) of the composite sample, $\lVert \cdot \lVert_{L_2}$ represents the L2 norm, and $\bar{\textbf{x}}= \{ \textbf{x}, t \}$. In the loss function, there are four different components play distinct roles: The first component quantifies the deviation between the model's rollout predictions and the experimental labels of mass across the entire temporal sequences; the second component acts as a regularization term, promoting sparsity; the third component seeks to impose trajectory smoothness; while the last loss component serves to ensure efficient convergence of the elliptic solver and also guide the model to avoid parameter regions where the elliptic solver may become unstable, stabilizing the training process. The coefficients $\beta_1$, $\beta_2$, and $\beta_3$ are regularization term weights, and their magnitudes are typically maintained at low values to preserve the fundamental nature of regularization. The objective of the PiNDiff model training is to minimize the total loss function $\mathcal{L}(\boldsymbol\theta)$, seeking the optimal parameter set $\boldsymbol{\theta}^*$ that most accurately aligns model predictions with experimental observations, subject to the described constraints and regularizations.

\subsection{Uncertainty quantification}

The PiNDiff I-CVI model fuses a physics-derived model, sparse measurements, and neural operators, aiming to capture the complex spatio-temporal behaviors of the I-CVI process. Yet, challenges arise from incomplete physics knowledge, the potential over-parameterization of DNNs, and the ever-present issue of data scarcity. These factors can compromise the model's prediction reliability, especially when the extent of missing physics is pronounced. In light of these challenges, UQ becomes a critical step, ensuring a more reliable and robust prediction framework for the I-CVI process.

In order to address the uncertainties inherent in PiNDiff model predictions based on training dataset $\mathcal{D}$, it is essential to determine the posterior distribution over the model parameters, denoted as $p(\boldsymbol{\theta} | \mathcal{D})$. Employing gradient descent leads to a singular realization of $p(\boldsymbol{\theta} | \mathcal{D})$, resulting in the predicted state $\textbf{v}(\bar{\textbf{x}}| \boldsymbol{\theta})$ being one realization of the model output, modeled as random variables $\textbf{V}(\bar{\textbf{x}}| \boldsymbol{\theta})$. Namely, the inherent model prediction uncertainty can be quantified by the probability distribution $p(\mathbf{V} | \bar{\mathbf{x}}, \mathcal{D})$, which can be obtained using the Bayesian Model Averaging (BMA),
\begin{equation}
    p(\mathbf{V}|\tilde{\mathbf{x}} , \mathcal{D}) = \int p(\mathbf{V} | \tilde{\mathbf{x}}, \boldsymbol{\theta}) p(\boldsymbol{\theta} | \mathcal{D}) d\boldsymbol{\theta},
    \label{eq:pred}
\end{equation}
and approximated using Monte Carlo integration,
\begin{equation}
    p(\mathbf{V}|\tilde{\mathbf{x}} , \mathcal{D}) \approx \frac{1}{M}\sum_{j=1}^M p(\mathbf{V} | \tilde{\mathbf{x}}, \boldsymbol{\theta}^{(j)}), \quad \boldsymbol{{\theta}}^{(j)} \sim p(\boldsymbol{\theta} | \mathcal{D}),
    \label{eq:pred_MC}
\end{equation}
where posterior distribution $p(\boldsymbol{\theta} | \mathcal{D})$ of trainable parameters is theoretically computed using Bayes' theorem, 
\begin{equation}
    p(\boldsymbol{\theta}|\mathcal{D}) = \frac{p(\mathcal{D|\boldsymbol{\theta}})p(\boldsymbol{\theta})}{\int p(\mathcal{D}|\boldsymbol{\theta}) p(\boldsymbol{\theta}) d\boldsymbol{\theta}}. 
    \label{eq:posterior}
\end{equation}  
In this context, $p(\boldsymbol{\theta})$ represents the prior distribution over $\boldsymbol{\theta}$ and $p(\mathcal{D|\boldsymbol{\theta}})$ is the joint likelihood of the dataset. Obtaining the posterior, as indicated by Eq.~\ref{eq:posterior}, either analytically or through traditional Bayesian sampling techniques, is computationally intractable~\cite{psaros2023uncertainty}. As a solution, we employ the DeepEn method~\cite{lakshminarayanan2017simple} here to tackle this challenge. The increasing appeal of ensemble-based techniques lies in their capacity to explore the multi-modal posterior landscape and their straightforward implementation~\cite{wen2020batchensemble}. To effectively approximate parameter distribution and gauge model uncertainty, multiple PiNDiff model instances are trained in parallel, each with a distinct initialization. This approach enables capturing multiple local maximum a posteriori (MAP) estimates in the posterior distribution for the parameter $\boldsymbol{\theta}$, each corresponding to distinct local minima within the landscape of the loss function $\mathcal{L}(\boldsymbol{\theta})$.

During the inference stage, predictions for each variable $\textbf{v}$ are derived using MAP samples $\{\boldsymbol{\theta}^{(j)}\}_{j=1}^M$ obtained through DeepEn training. As a result, the mean and variance of the model predictions are calculated as,
\begin{subequations}
    \begin{alignat}{2}
    \mathbb{E}(\textbf{V}|\bar{\textbf{x}}, \mathcal{D}) &\approx \frac{1}{M} \sum^M_{j=1} \textbf{v}(\bar{\textbf{x}} |\boldsymbol{\theta}^{(j)}),\\
    \mathbb{V}\mathrm{ar}(\textbf{V}|\bar{\textbf{x}}, \mathcal{D}) &\approx  \frac{1}{M} \sum^M_{j=1} (\textbf{v}(\bar{\textbf{x}}|\boldsymbol{\theta}^{(j)}) - \mathbb{E}(\textbf{V}|\bar{\textbf{x}}, \mathcal{D}))^2 .
    \end{alignat}
    \label{eq:pred_approx}
\end{subequations}
In this context, $\mathbb{E}(\textbf{V}|\bar{\textbf{x}}, \mathcal{D})$ is the finalized model prediction, and thrice the value of variance $\mathbb{V}\mathrm{ar}(\textbf{V}|\bar{\textbf{x}}, \mathcal{D})$ (considering only its diagonal component) is used to define the confidence interval. These estimations of mean and variance equip us with a more robust predictive capability and facilitate quantifying the uncertainties inherent in the model's predictions.

\subsubsection{Synthetic data generation}
\label{sec:syn-data-gen}
To assess the effectiveness of the proposed I-CVI PiNDiff model, a series of virtual experiments have been conducted as preliminary steps prior to using the real-world data. This necessitates the generation of synthetic data for model training, where ground truth is available for validation. To this end, a physics-based numerical model is developed, drawing inspiration from the formulation presented by Wei et al.~\cite{wei2006two}. Gaussian noise is added to the synthetic data to emulate real-world conditions. Specifically, the general model of C/SiC deposition for synthetic data generation is based on Eq.~\ref{eq:I-CVI-simp}. For the carbon deposition in the I-CVI cycle, the ``true'' parameter values for data generation are defined as follows: $M_d=0.01199 , \text{kgmol}^{-1}$, $\rho_d=2260 , \text{kgm}^{-3}$, $q=1$, and $K=k_0\exp\left(\frac{-E_r}{RT}\right)$ with $k_0=2.62 \text{(m/s)}$ and $E_r=1.46 \times 10^5 , \text{Jmol}^{-1}$.
The effective diffusion coefficient is determined as $D_{eff} = \frac{\varepsilon D_KD_{AB}}{\tau_0 (D_K+D_{AB})}$, where
$D_{AB}$ is the binary diffusion coefficient, $\tau_0$ is the torosity, and the Knudsen diffusion coefficient is given as $D_K = \frac{2}{3} \left(\frac{8RT}{\pi M}\right)^{1/2} r$; $r$ represents the characteristic pore radius in meters. For the purpose of demonstration, the binary diffusion coefficient is simplified as $D_{AB} = 10^{-5} \frac{T^{1.75}}{P}$ and torosity as $\tau_0=6.78$. The true model form of surface-to-volume ratio $S_v$ is defined as,
\begin{equation}
    S_{v, true}  = \frac{\varepsilon}{\varepsilon_0} \frac{1-\varepsilon_0}{r_f} \Big[ 1-\Big( \frac{\varepsilon_0}{1-\varepsilon_0} \Big) ln\Big(\frac{\varepsilon}{\varepsilon_0} \Big) \Big]^{1/2},
    \label{eq:trueS_v}
\end{equation}
where $\varepsilon_0$ is the initial porosity, and for synthetic data generation, $\varepsilon_0=0.6$.
Given that the C/C composite preforms examined in this investigation exhibit axisymmetry, our simulations are performed within a 2-dimensional axisymmetric domain, assuming no variation of physical properties in the azimuthal plane. The resulting global variables, such as mass, are obtained by integrating over the entire 3-dimensional domain.

\section{Results}
\label{sec:result}

The PiNDiff I-CVI model, enhanced with a probabilistic learning approach, is evaluated using both synthetic and real experiment datasets from C/C composite manufacturing processes. With the synthetic dataset, a comprehensive set of numerical experiments is conducted to investigate the model's performance in inference, prediction accuracy, and its capability to quantify uncertainty across diverse training settings. In evaluating its real-world applicability, the model also undergoes validation with an experimental dataset sourced from Benzinger and Huttinger's work~\cite{benzinger1998chemical, benzinger1999chemical}. This exercise allows an exploration of the model's effectiveness in forecasting the densification process, especially under varied partial pressures, some of which deviate from those seen during model training. Extending the model's evaluative scope, the PiNDiff approach is employed to model the scenarios involving multiple I-CVI cycles, mirroring the manufacturing protocols employed by Honeywell Inc. when producing C/C composites for airplane brake systems. Both synthetic and real multi-CVI datasets from Honeywell are used for this assessment.

\subsection{Inference of unknown physics}
\begin{figure}[t!]
\centering
\includegraphics[width = 1.0\textwidth]{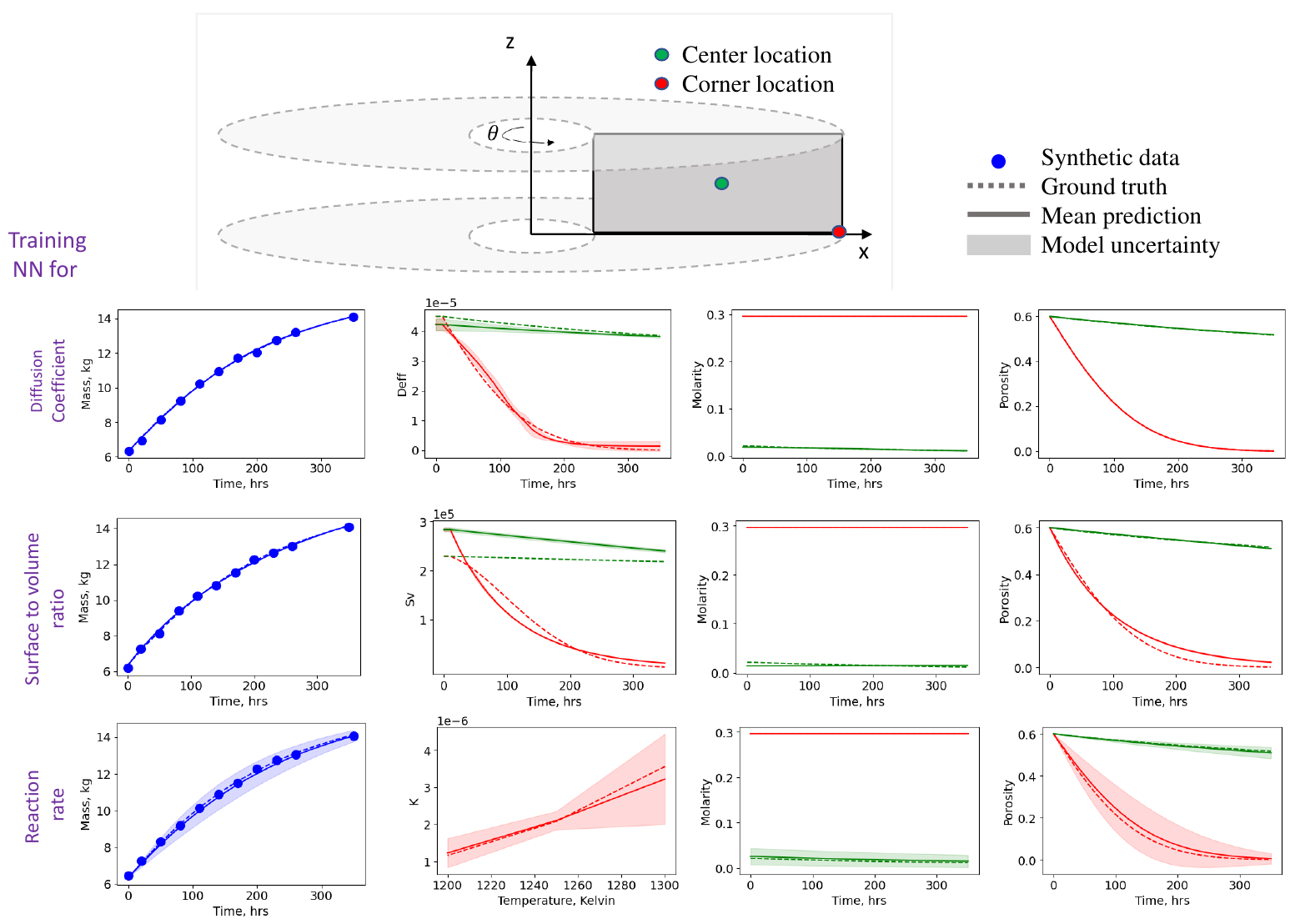}
\caption{Prediction and inference results of the trained PiNDiff I-CVI model with quantified uncertainty. The operators $D_{eff}$, $K$, and $S_v$ are trained independently.}
\label{fig:Indpdnt_NN_train}
\end{figure}

The PiNDiff I-CVI model incorporates three different neural networks, $\mathcal{D}_{nn}(\cdot;\boldsymbol{\theta}_{D_{eff}})$, $\mathcal{K}_{nn}(\cdot;\boldsymbol{\theta}_K)$, and $\mathcal{S}_{nn}(\cdot;\boldsymbol{\theta}_{S_v})$, to capture the unknown operators $D_{eff}(\mathbf{x}, t)$, $K(\mathbf{x}, t)$, and $S_v(\mathbf{x}, t)$, respectively, as detailed in Section \ref{sec:methodology}. To evaluate the model's capability in inferring these unknown fields from indirect mass measurements, we conducted a set of three experiments using synthetic data. In each experiment, one of the operators --either $D_{eff}$, $K$, or $S_v$ -- was treated as unknown and to be inferred with its corresponding neural network. The representations for the other two operators were provided based on the ``ground truth'' model presented in section \ref{sec:syn-data-gen}. Due to the inherent difficulties in collecting experimental data, the training dataset exclusively comprises measurements of the preform's mass taken at varied time intervals throughout the I-CVI cycle. Namely, only these sparse mass data points are used for model training (blue dots in the first column of Fig.~\ref{fig:Indpdnt_NN_train}).
In Fig~\ref{fig:Indpdnt_NN_train}, the PiNDiff model's predicted mass growth (represented by the solid line) aligns well with the synthetic data (indicated by dots) across all three cases. The trained model is also able to accurately capture the spatio-temporal dynamics of both the unknown parameters ($D_{eff}$, $S_v$, and $K$) and state variables (porosity and molarity), as evident by examining two critical locations on the preform: it's center (green) and corner or surface (red). The close match between the model's predicted mean (solid line) and the ground truth (dashed line) underscores the PiNDiff model's capability to infer unobserved states and parameters from limited, indirect observations. Furthermore, the model provides a quantification of the uncertainty associated with each prediction. These uncertainties are notably low, given that only one of the functional representations is assumed to be unknown.    

\begin{figure}[t!]
\centering
\includegraphics[width = 1.0\textwidth]{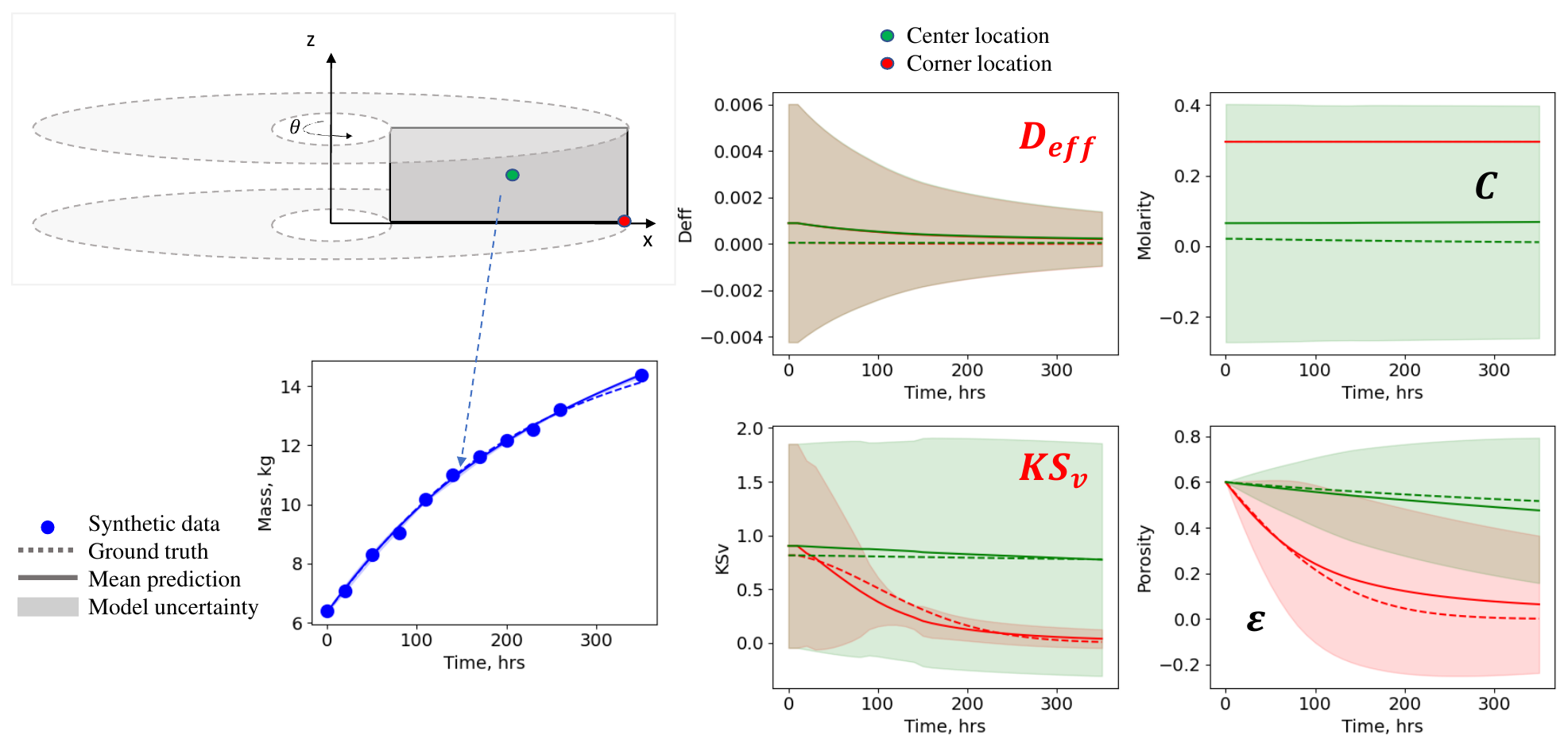}
\caption{Prediction and inference results of the trained PiNDiff I-CVI model with quantified uncertainty. All three operators $D_{eff}$, $K$, and $S_v$ are trained simultaneously.}
\label{fig:All_NN_train}
\end{figure}
An extended investigation was conducted where the functional forms of all three operators, $D_{eff}$, $K$, and $S_v$, were treated as unknown and trained using the same dataset. The results of this exploration are illustrated in Fig.~\ref{fig:All_NN_train}. Although the predictions for mass remain accurate, the inferred mean values for the hidden states and parameters are slightly deviate more than those observed in previous experiments. This discrepancy arises due to the increased complexity introduced by a diminished extent of known physics. However, the model's uncertainty in its predictions is elevated, indicating a reduced confidence in its predictions. This highlights the PiNDiff model's ability to reasonably signal prediction confidence through UQ when faced with different levels of unknown physics of the system.  Moving forward, all subsequent analyses will operate under the assumption that $D_{eff}$, $K$, and $S_v$ are unknown, mirroring real-world scenarios.

\subsection{Generalizability with respect to operating conditions}
To assess the model's generalizability across different input parameters, we trained it using synthetic data generated from a diverse range of operating conditions. Subsequently, it was validated against different sets of operating conditions not encountered during training. This approach provides insight into the model's capability to handle and predict the I-CVI process for new input scenarios beyond its training conditions.

\subsubsection{Case 1: Training using segmented sample mass data}

\begin{figure}[ht]
\centering
\includegraphics[width = 1.0\textwidth]{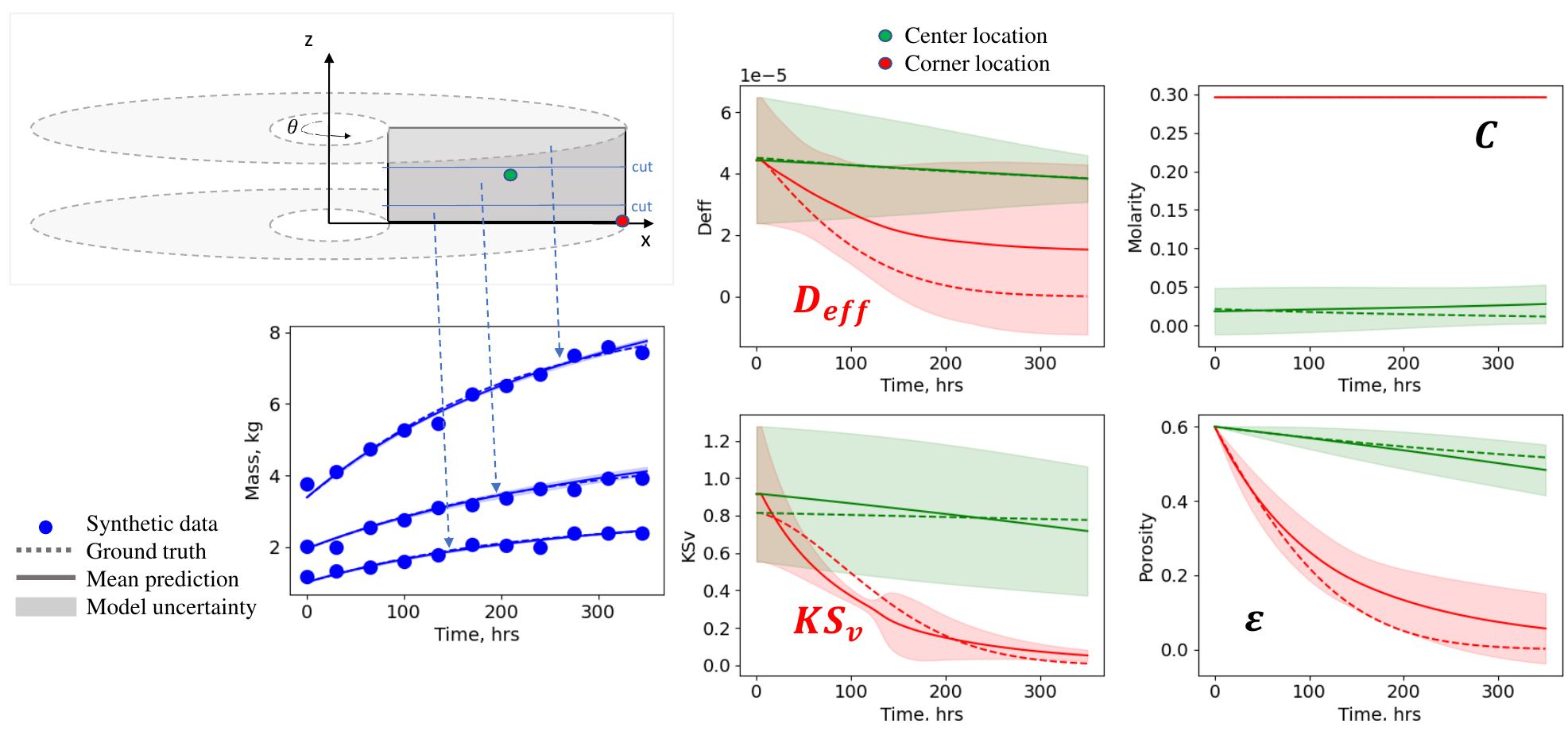}
\caption{Prediction and inference results of the trained PiNDiff I-CVI model with quantified uncertainty. The model is trained with synthetic data of mass for three pieces.}
\label{fig:Pred_3_cut_synthetic}
\end{figure}

\begin{figure}[!t]
\centering
\includegraphics[width = 1.0\textwidth]{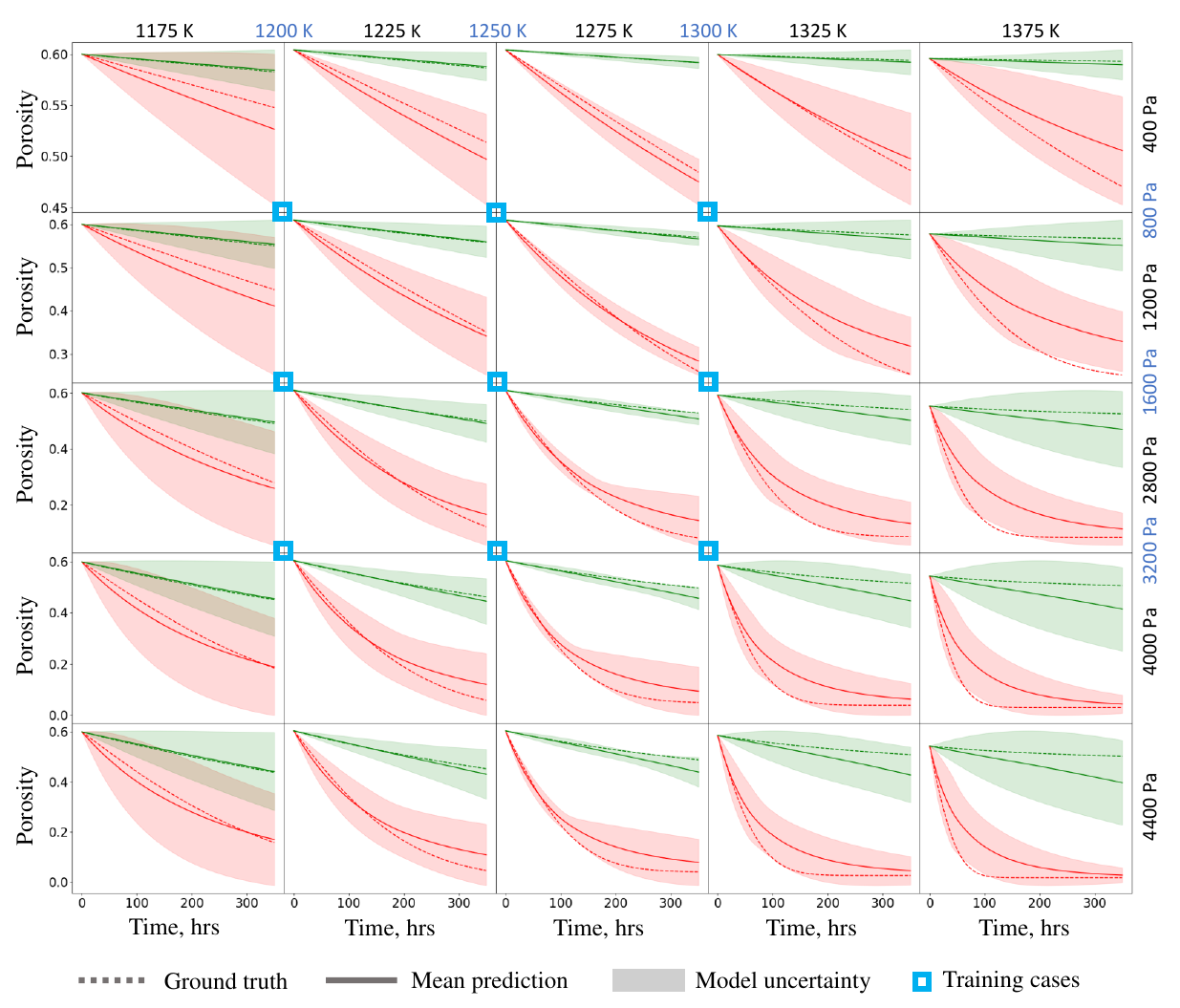}
\caption{PiNDiff predictions on porosity with uncertainty for unseen partial pressure and temperature. The model is trained on synthetic mass data for three pieces.}
\label{fig:Pred_3_cut_synthetic_test}
\end{figure}

To facilitate the I-CVI model in learning spatial dynamics, the preform sample was segmented into three non-uniform sections along the z-direction, as visualized in the top-left panel of Fig.~\ref{fig:Pred_3_cut_synthetic}. The training was performed using mass data of these segmented sections over time, as shown by blue dots. These noisy data were generated under nine varying conditions, involving three different temperatures (1200 K, 1250 K, and 1300 K) and three different partial pressures (800 Pa, 1600 Pa, and 3200 Pa), as illustrated in Fig.~\ref{fig:Pred_3_cut_synthetic_test}.
The model's excellent predictive capability is evident in the bottom-left segment of Fig.~\ref{fig:Pred_3_cut_synthetic}, where its estimations closely match the projected mass evolution across a span of 350 hours for each piece. Additionally, the model adeptily infers the spatio-temporal trajectories of porosity $\varepsilon$, molarity $C$, effective diffusion coefficient $D_{eff}$, and effective deposition rate $KSv$. As shown in Fig.~\ref{fig:Pred_3_cut_synthetic}, the confidence interval provided by the model adequately encompasses the ground truth for the entire duration, indicating that the proposed PiNDiff model can effectively uncover the hidden states/parameters from indirect data with quantified uncertainty.

To assess the generalizability of the trained I-CVI model, we test it to a range of unseen operating conditions, spanning different temperatures $T$ and partial pressures $P$. This evaluation covers both interpolation and extrapolation domains within this $T$-$P$ parameter space. The results, particularly the predicted porosity across these testing conditions, are visualized in Fig.~\ref{fig:Pred_3_cut_synthetic_test}. For reference, conditions encountered during training are marked by blue squares within the $T$-$P$ space. Note that the dynamics of predicted porosity at the two distinct locations are displayed solely for the testing conditions. For the operating conditions within the interpolation domain, the model demonstrates a higher level of accuracy -- its predictions aligned well with the ground truth and are accompanied by a relatively confined uncertainty range. Remarkably, even though the model's training was based on the dataset obtained from just nine operating conditions (blue squares),  it still demonstrates strong capability in extrapolation. As shown in Fig.~\ref{fig:Pred_3_cut_synthetic_test}, for the temperatures and partial pressures beyond the training range, the model not only delivers accurate predictions but also maintains a credible uncertainty envelope that encompasses the ground truth. As we move further from the training conditions, the confidence intervals of model predictions increase, indicating a higher uncertainty. It is clear that the extrapolative predictions are naturally paired with a more expansive uncertainty envelope compared to those from the interpolation domain, demonstrating that the model is able to account for potentially diminished prediction fidelity in these unfamiliar zones.

\subsubsection{Case 2: Training using total mass data of entire sample}

\begin{figure}[!t]
\centering
\includegraphics[width = 1.0\textwidth]{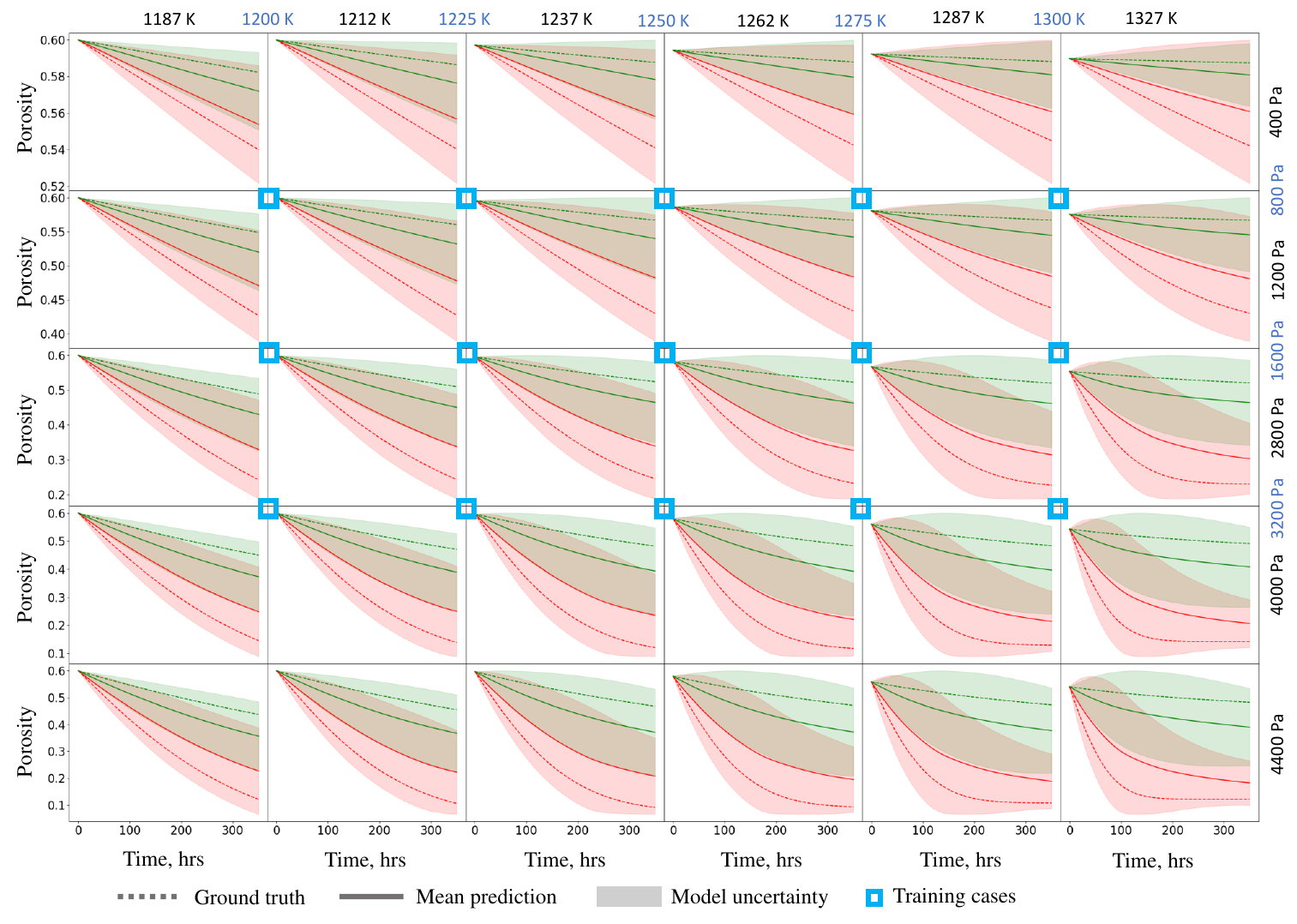}
\caption{PiNDiff predictions on porosity with uncertainty for unseen partial pressure and temperature. The model is trained on synthetic total mass data.}
\label{fig:Pred_1_cut_synthetic_test}
\end{figure}

Obtaining detailed measurement data by segmenting multiple C/C composite samples during the manufacturing process can be labor-intensive and often impractical. Building upon our earlier study, here we focus on assessing the model's ability to generalize within the input operating parameter space using only the total mass data from the entire unsegmented composite sample. This implies that our training dataset lacks detailed spatial information, presenting a substantial challenge to the model's capacity for spatial field inference and prediction. Specifically, the training of the PiNDiff model, in this case, is exclusively based on total mass data. To enhance the model's training, a richer set of conditions was included, including five temperatures (1200 K, 1225 K, 1250 K, 1275 K, and 1300 K) and three partial pressures (800 Pa, 1600 Pa, and 3200 Pa), leading to a total of 18 training conditions, as illustrated by blue squares in Fig.~\ref{fig:Pred_1_cut_synthetic_test}. This broader spectrum was deemed necessary due to the absence of spatial information in the training set.
\begin{figure}[ht]
\centering
\includegraphics[width = 1.0\textwidth]{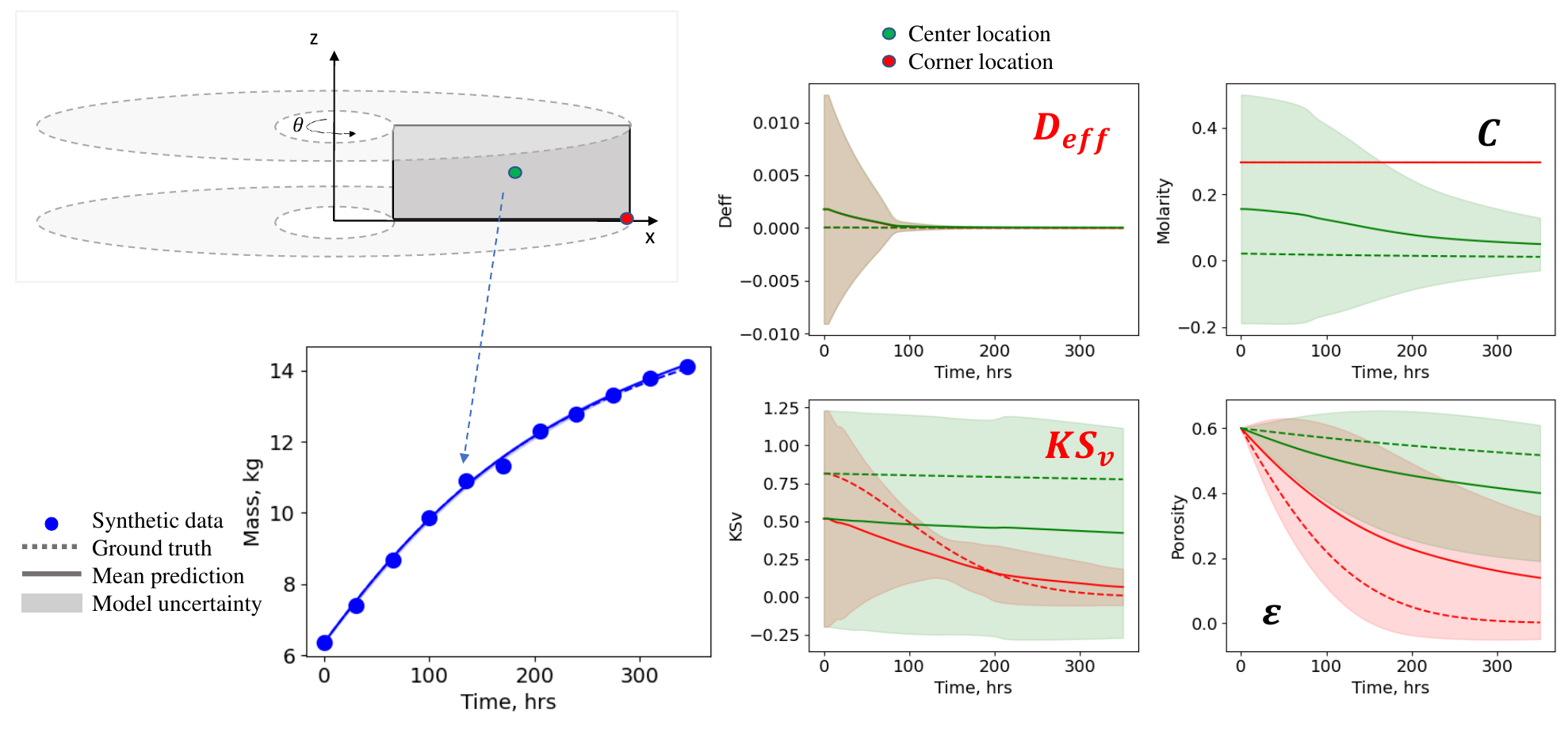}
\caption{Prediction and inference results of the trained PiNDiff I-CVI model with quantified uncertainty. Model is trained with synthetic data of total mass for the entire composite sample.}
\label{fig:Pred_1_cut_synthetic}
\end{figure} 
The prediction and inference results of one training condition are shown in Fig.~\ref{fig:Pred_1_cut_synthetic}. As displayed in the bottom-left of Fig.~\ref{fig:Pred_1_cut_synthetic}, the trained model can accurately trace the mass trajectory over a 350-hour interval with minimum uncertainty. However, the inferred spatio-temporal dynamics for $D_{eff}$ and $KSv$ are notably deviated from the ground truth, primarily attributed to the limited and spatially non-informative data. Nonetheless, the estimated uncertainty of the prediction signals its inaccuracies, and the confidence interval still envelopes the ground truth. Despite the less accurate inferences for $D_{eff}$ and $KS_v$, the spatio-temporal behavior of the porosity and molarity remains plausible, which can be credited to the PiNDiff model's inherent capability of maintaining the mathematical structure of the physics-derived models. Furthermore, the model's uncertainty predictions for molarity and porosity are both reasonable and inclusive of the ground truth. Hence, even when faced with limited and less descriptive data, the model remains resilient, with its quantified uncertainty adeptly capturing variations in data quality. 
The PiNDiff model, once trained, is tested under operating conditions that were not part of the training set, spanning both interpolation and extrapolation regions within the $T$-$P$ parameter space. Fig.~\ref{fig:Pred_1_cut_synthetic_test} presents the results for the predicted porosity under these unseen conditions. Similar to the previous experiment, the model continues to provide reliable spatio-temporal porosity predictions for both interpolated and extrapolated unseen operating conditions, and its uncertainty reasonably covers the ground truth. Compared to the previous cases with detailed mass data of segmented samples, the mean prediction of porosity in this scenario tends to be less accurate and the uncertainty margin also increases when training is reliant on total mass measurements of the entire sample.

\subsection{Training on real experimental data from literature}
Previous studies primarily evaluate the PiNDiff I-CVI model using synthetic datasets. In this section, we shift the focus to its performance on real-world data. In particular, the training and testing of the PiNDiff model are based on real experimental data, as detailed in Benzinger and Huttinger's work~\cite{benzinger1998chemical, benzinger1999chemical}. Their seminal experiments concentrated on the I-CVI densification process, maintaining methane at a constant temperature of $1100^o C$, while adjusting total pressures to either 20 kPa~\cite{benzinger1998chemical} or 100 kPa\cite{benzinger1999chemical}. Within these setups, methane's partial pressure was modulated and varied during the experiments. The selected substrate for these experiments was a cylindrical, porous alumina ceramic, with dimensions of 20 mm in height and 16 mm in diameter and a total porosity of 23\%. 
\begin{figure}[!ht]
\centering
\includegraphics[width = 0.8\textwidth]{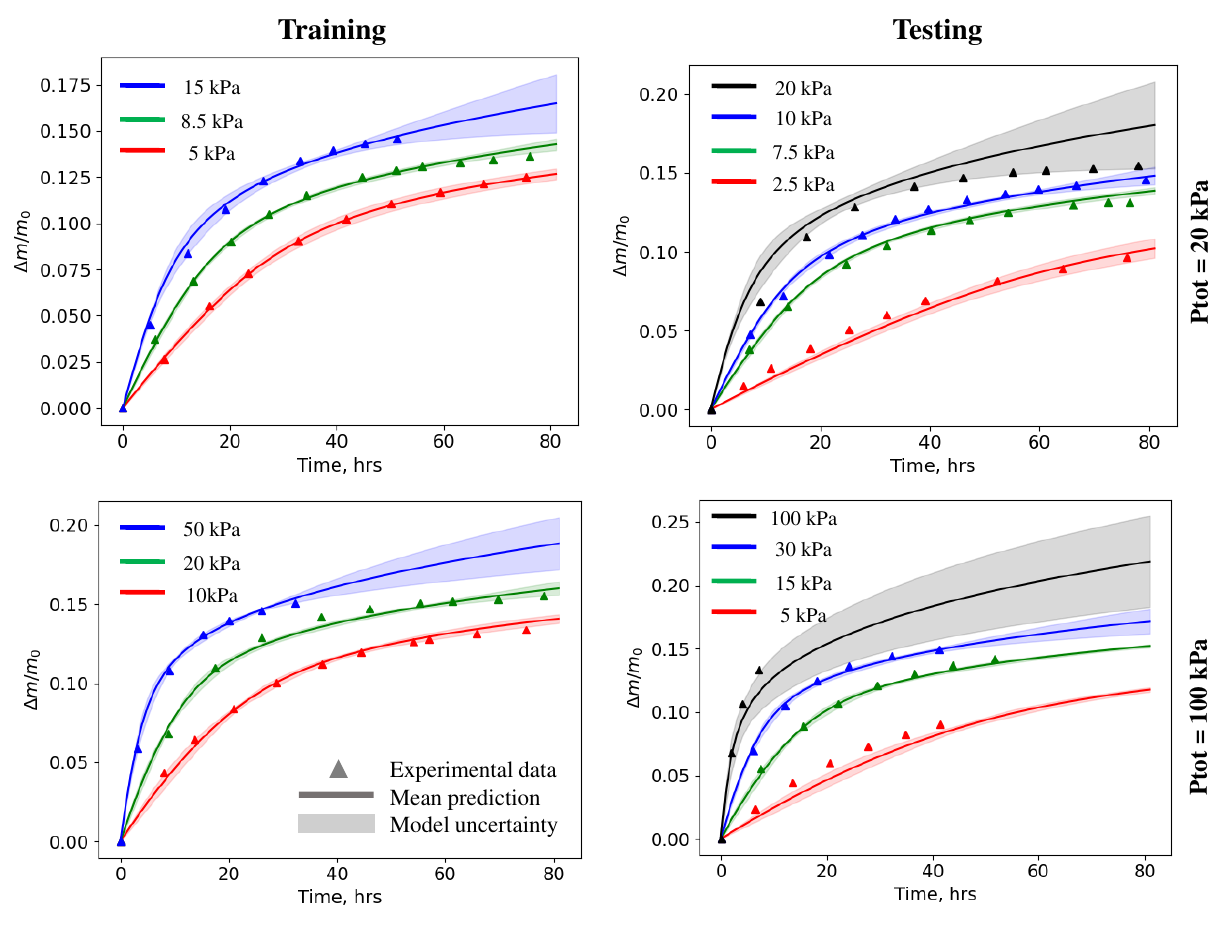}
\caption{PiNDiff predictions with uncertainty on Benzinger and Huttinger's experimental mass data at various partial pressure~\cite{benzinger1998chemical, benzinger1999chemical}.}
\label{fig:BenExpComp}
\end{figure}
During the experiment, the mass gain over time is measured given different partial pressures, and thus, two datasets at different total pressures are available. These sparse mass measurement data are plotted as triangles in Fig.~\ref{fig:BenExpComp}. For both cases, the PiNDiff I-CVI model is trained with mass data from three different partial pressures and subsequently undergoes testing on the other four partial pressures absent from the training set. Notably, two of these test scenarios were in the extrapolation zone, falling outside the training condition range. Separate trainings are carried out for each experimental dataset, corresponding to the 20 kPa and 100 kPa total pressures. 
Comparative analysis between the PiNDiff I-CVI model's predictions and the real measurements across both training and testing scenarios can be viewed in Fig.~\ref{fig:BenExpComp}. The model's predictions exhibited excellent alignment with the real data for both training and testing partial pressures, attesting the learning and predictive capability of the proposed model. Moreover, it is important to note that the model's uncertainty predictions are higher in regions where the training data are absent. And, during testing phases, higher prediction uncertainty is observed for elevated pressure values, suggesting diminished confidence in the model's predictions in those specific regions.

\subsection{Modeling densification process via multiple I-CVI cycles}

Previous results have demonstrated the PiNDiff I-CVI model's proficiency in learning unknown physics operators across various operating parameters. Consequently, the model holds great potential for predictive and optimization tasks. In industrial manufacturing settings, however, gathering experimental data under varied operational conditions can be prohibitively costly and time-intensive. The routine extraction of composites for mass data collection throughout the process also poses practical challenges. Therefore, in large-scale applications, such constraints might result in very limited data across the parameter space, thereby restricting the model's training and predictive capabilities under different conditions. However, with substantial knowledge—like the porosity field derived from the PiNDiff I-CVI model—it becomes feasible to postulate parameters for subsequent experiments, facilitating optimization. In this section, we focus on evaluating the model's inference capability in a more realistic manufacturing setting but under a single operational condition.

\begin{figure}[!t]
\centering
\includegraphics[width = 0.8\textwidth]{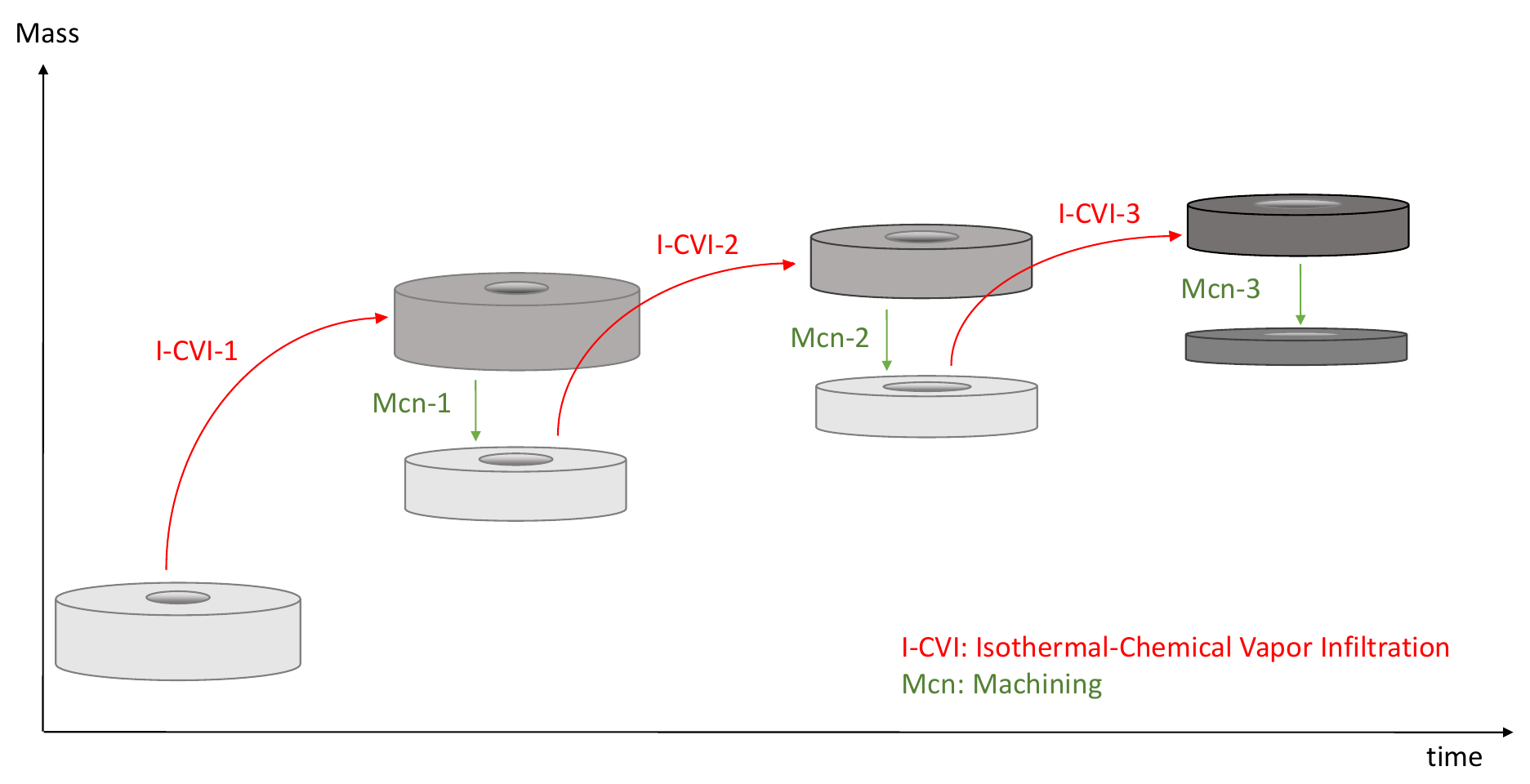}
\caption{Multi-I-CVI manufacturing process with intermediate machining.}
\label{fig:HW_CVI_process}
\end{figure}
During the I-CVI process, the concentration of the precursor reactant gas diminishes as it diffuses deeper into the preform, leading to a decelerated deposition rate towards the preform's core. As a result, deposition on the preform's external surface is faster compared to its interior. This preferential exterior deposition causes the outer pores to seal prematurely before the interior is adequately filled, obstructing subsequent gas diffusion and impeding the densification process. In response to this challenge, industrial entities such as Honeywell have adopted a technique of machining that removes the outer layer to expose these pores inside again. Consequently, we need a series of multiple I-CVI cycles performed consecutively, along with the machining process, to ensure enhanced densification. Figure~\ref{fig:HW_CVI_process} provides a schematic representation of three such interlinked I-CVI and machining cycles. This cyclic methodology is adopted by Honeywell in their C/C composite brake manufacturing protocol. Typically, during the I-CVI phases, the brakes gain mass, which is then reduced during the machining stages. Data harvested from Honeywell's iterative I-CVI brake production serves as the training data for our PiNDiff I-CVI model in inferring porosity distribution. Nevertheless, a challenge arises in Honeywell's real manufacturing cases as the inferred quantities lack accompanying validation data. To address this, initial cross-validation of the model is undertaken using synthetic data of multi-CVI/machining cycles, synthesized in alignment with established industry practices.

The proposed PiNDiff I-CVI model is modified to accommodate multiple I-CVI cycles interspersed with machining phases. In the machining interval, attributes like geometry, computational grid, and foundational physical parameters (e.g., porosity from the previous cycle) are re-calibrated in line with the post-machining preform specifications. Synthetic mass data is collected at the end of each cycle.

\begin{figure}[!t]
\centering
\includegraphics[width = 1.0\textwidth]{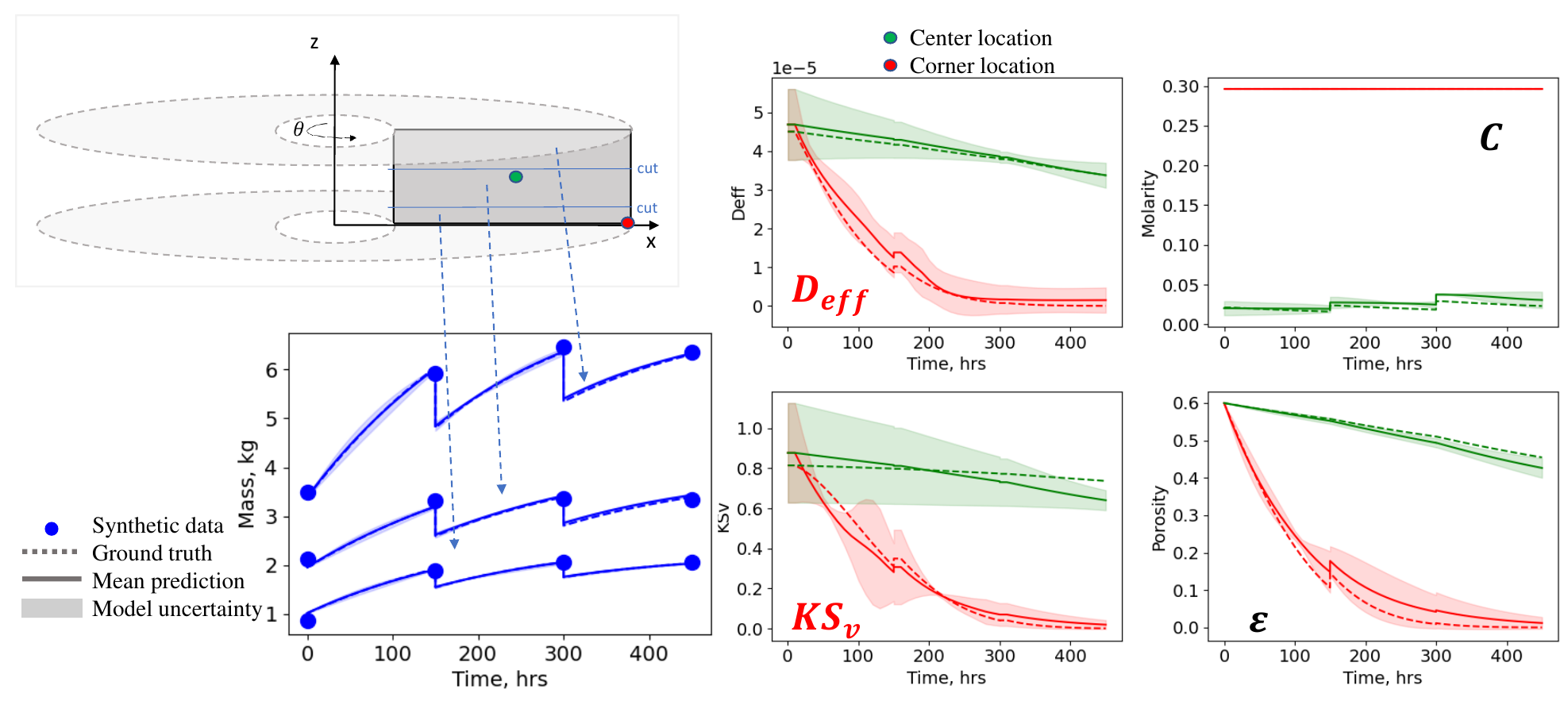}
\caption{Prediction and inference results of the trained PiNDiff model with quantified uncertainty for multi-I-CVI and machining cycles. Model is trained with synthetic data of mass for three pieces.}
\label{fig:pred_3c3x}
\end{figure}
Initially, the model is trained using synthetic data, aiming to assess its inference capability for a given operating condition. The data used for training include mass measurements of three unevenly segmented portions cut along the z-axis of the brake sample (blue dots in Fig.~\ref{fig:pred_3c3x}). After undergoing 500 epochs of training, the PiNDiff model can accurately forecast the incremental brake mass gain throughout the manufacturing cycles, as illustrated in Fig.~\ref{fig:pred_3c3x}. Moreover, the model also accurately infers the spatio-temporal dynamics of key hidden parameters and states such as $D_{eff}$, $KSv$, molarity, and porosity, with its predictions aligning closely with the ground truth, as evinced by their confinement within the confidence intervals. In line with the previous results, the uncertainty is lower as the training data contains spatial information. While the model's predictions might warrant caution under varied operational scenarios, the derived porosity remains crucial, offering substantive insights into the composite's characteristics, and thereby facilitating the subsequent optimization endeavors. 

\begin{figure}[!t]
\centering
\includegraphics[width = 1.0\textwidth]{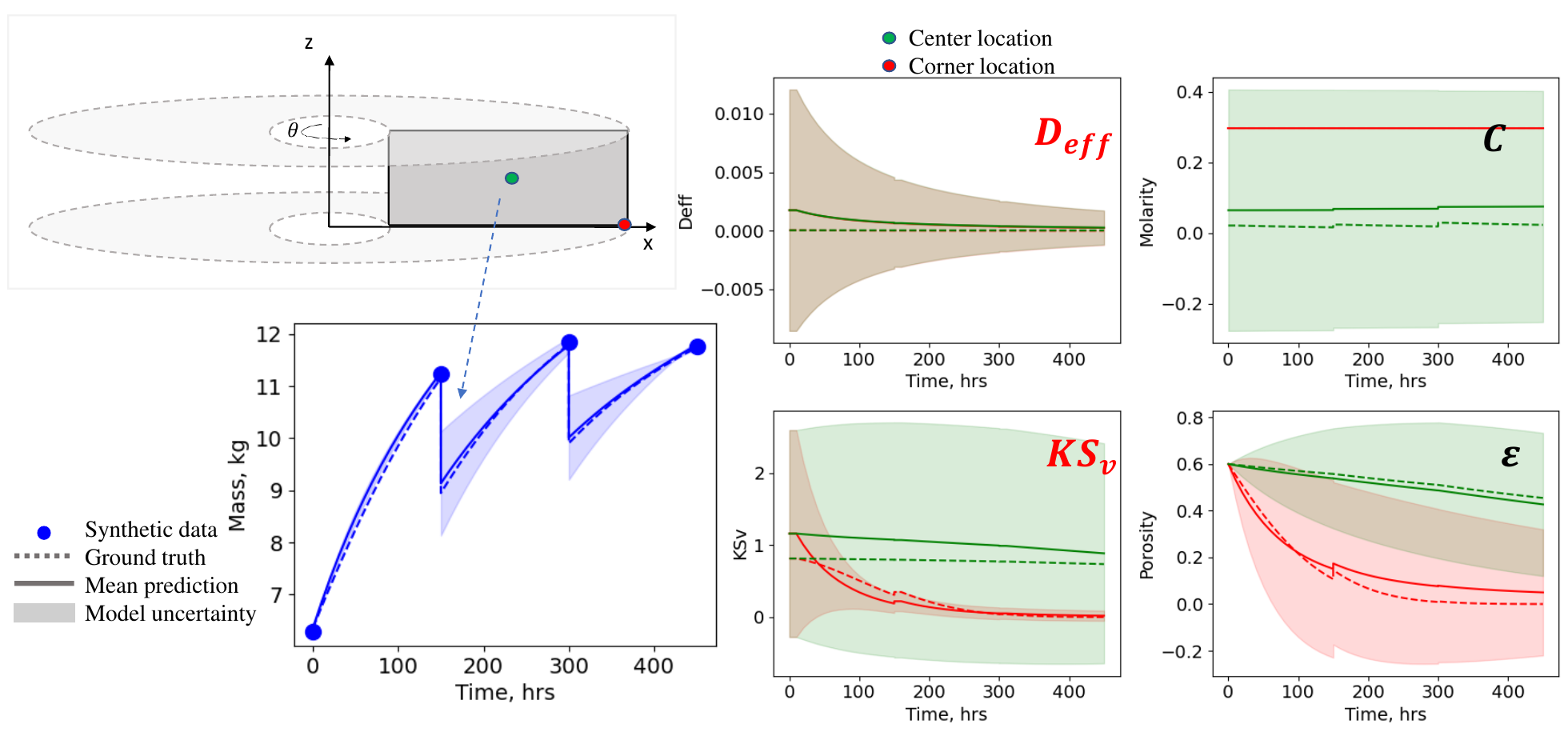}
\caption{Prediction and inference results of the trained PiNDiff model with quantified uncertainty for multi-I-CVI and machining cycles. The model is trained with synthetic data of the total mass for an entire brake.}
\label{fig:pred_3c1x}
\end{figure}
The PiNDiff model's efficacy is subsequently examined using the total mass as synthetic data rather than multiple cuts. Illustrated in Fig.~\ref{fig:pred_3c1x}, the results follow a similar trend to the previous case. After a training of 500 epochs, the model accurately predicts the mass trajectory throughout the fabrication process. Additionally, the hidden parameters and states like $D_{eff}$, $KSv$, molarity, and porosity are also estimated by the model, where the mean values agree with the ground truth reasonably well. It's noteworthy that, in this setup, the model's uncertainty bounds are more expansive than those in the previous scenario that utilizes localized mass data. The increased uncertainty arises from the comparatively diminished spatial information in the present training dataset.    

\begin{figure}[!t]
\centering
\includegraphics[width = 1.0\textwidth]{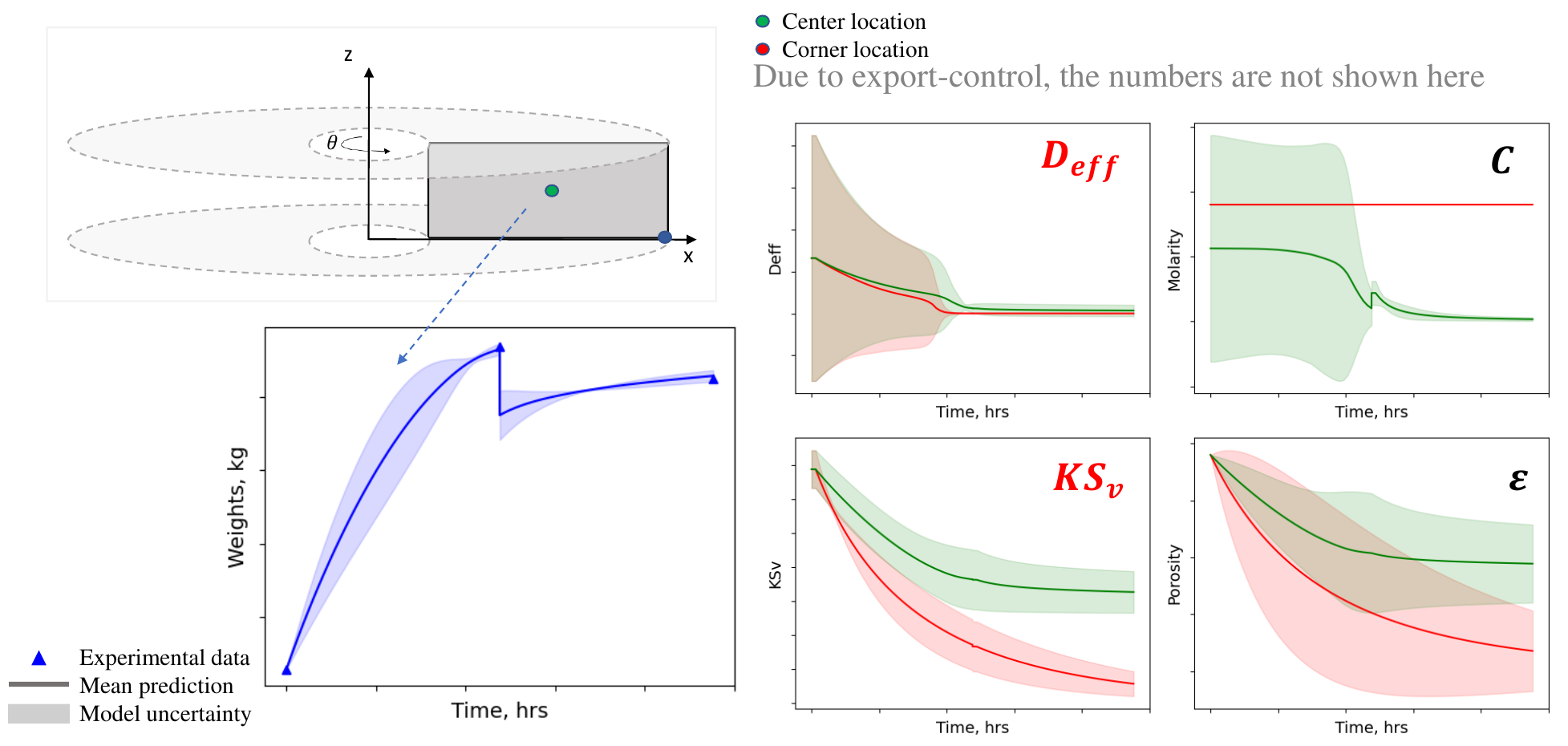}
\caption{Prediction and inference results of the trained PiNDiff model with quantified uncertainty, based on the Honeywell manufacturing dataset. (Specific scales have been omitted due to export-control restrictions). }
\label{fig:pred_HW}
\end{figure}

Lastly, we evaluate the PiNDiff model for learning and predicting multiple CVI cycles using the real measurement data from Honeywell's C/C brake manufacturing processes. The results are presented in Fig.~\ref{fig:pred_HW}. The trained model successfully captures the evolution of the densification process throughout the manufacturing phase. Crucially, during its training, the model offers spatio-temporal forecasts of porosity, an indispensable parameter for optimization endeavors. While no supplementary experimental data exists for direct validation of inferred quantities, the model's performance with synthetic data suggests a substantial degree of reliability in its predictions. Such capabilities of the PiNDiff I-CVI model hold significant potential for enhancing optimization strategies of CVI processes and informed decision-making in the industry.

\section{Conclusion}
\label{sec:conclusion}

This study presents the development and assessment of a physics-integrated neural differentiable (PiNDiff) model for I-CVI processes. By seamlessly integrating the partially known physics into a deep learning framework via differentiable programming, the PiNDiff model has established itself as an effective modeling framework for the I-CVI process, showing competency even when reliant on limited, indirect training datasets.

The PiNDiff model has been thoroughly evaluated through a series of experiments and tests, confirming its effectiveness and capabilities. Even when trained on limited and indirect synthetic data, the model exhibits promising performance, accurately predicting mass deposition and inferring the spatio-temporal fields of porosity, diffusion coefficient, deposition rate, and molarity. A salient feature of the model is its intrinsic capability to encapsulate and quantify uncertainties, particularly when confronted with incomplete physics and scarce data. Furthermore, the PiNDiff model demonstrates its capability to generalize across different operational conditions by accurately predicting the spatio-temporal dynamics of porosity under unseen temperatures and partial pressures. The model exhibits excellent performance in both interpolation and extrapolation scenarios, with lower model uncertainty observed in interpolation cases and higher uncertainty in extrapolation cases. This showcases the model's versatility and potential to effectively predict the I-CVI process under a wide range of conditions while accounting for its prediction confidence.

By leveraging experimental data, notably from studies by Benzinger and Huttinger, along with data obtained from Honeywell's manufacturing processes, the PiNDiff model reaffirmed its effectiveness and ability in predictive modeling of the I-CVI processes. Notably, the model successfully captures the higher uncertainty in regions where data was absent during training. The results highlight the model's potential to support process optimization and decision-making tasks, emphasizing its practical applicability in real-world scenarios. 

To conclude, the PiNDiff model proves to be a valuable tool for understanding, simulating, and predicting the I-CVI densification process. Its ability to integrate partially-known physics with deep learning enables effective predictive modeling using sparse and indirect measurements. While further comprehensive validation with more experimental data is desirable, the PiNDiff I-CVI model offers a promising approach for advancing the manufacturing of carbon-based composites to enhance their performance and quality.

\section*{Acknowledgment}
The authors would like to acknowledge the funds from the Air Force Office of Scientific Research (AFOSR), United States of America, under award number FA9550-22-1-0065. JXW would also like to acknowledge the funding support from the Office of Naval Research under award number N00014-23-1-2071 and the National Science Foundation under award number OAC-2047127 in supporting this study.

\section*{Compliance with Ethical Standards}
Conflict of Interest: The authors declare that they have no conflict of interest.




\bibliographystyle{elsarticle-num}
\bibliography{ref, bibtex_jwPub}

\clearpage
\appendix

\section{Nomenclature}

\begin{table}[!h]
    \centering
    \begin{tabular}{c c}
       $\varepsilon_k$  &  volume fraction of species $k$ \\
       $\varepsilon_g$  &  volume fraction of gas \\
       $\varepsilon$  &  porosity \\
       $\rho_k$  &  density of species $k$ [$kg/m^3$]\\
       $\textbf{u}_k$  &  velocity vector of species $k$ [$m/s$]\\
       $\dot{\omega}_k$  &  production rate of species $k$ [$kg/m^3s$]\\
       $D_{k,eff}$  &  effective diffusion coefficient of species $k$ [$kg/m^3s$]\\
       $C$  &  Morality of gas [$mol/m^3$]\\
       $M$  &  Molar mass of depositing species [$mol/m^3$]\\
       $S_v$  &  Effective infiltration area [$m^2/m^3$]\\
       $*_g$  &  gas\\
       $*_s$  &  solid\\
         & 
    \end{tabular}
    \caption{Caption}
    \label{tab:my_label}
\end{table}

\section{Derivation}
\label{sec:app-derivation}

The governing equations to simulate the I-CVI process for each species are given as \cite{mcallister1993simulation}
\begin{equation}
    \frac{\partial(\varepsilon_k\rho_k)}{\partial t} + \nabla.(\varepsilon_k\rho_k\textbf{u}_k) = D_{k} \nabla^2(\varepsilon_k\rho_k) + \dot{\omega}_k \ \ k=1,...,N
\end{equation}
\begin{subequations}
Gas
\begin{equation}
    \frac{\partial(\varepsilon_k\rho_k)}{\partial t} + \nabla.(\varepsilon_k\rho_k\textbf{u}_k) = D_{k} \nabla^2(\varepsilon_k\rho_k) + \dot{\omega}_k \ \ k=1,...,N_g
\end{equation}
Solid
\begin{equation}
    \frac{\partial(\varepsilon_k\rho_k)}{\partial t} + \nabla.(\varepsilon_k\rho_k\textbf{u}_k) = D_{k} \nabla^2(\varepsilon_k\rho_k) + \dot{\omega}_k \ \ k=1,...,N_s
\end{equation}
\end{subequations}

Summing over all gaseous and solid species results in\\
\begin{subequations}

\begin{equation}
    \frac{\partial(\varepsilon_g\rho_g)}{\partial t} + \nabla.(\varepsilon_g\rho_g\textbf{u}_g) = D_{eff,g} \nabla^2(\varepsilon_g\rho_g) + \dot{\omega}_g
\end{equation}

\begin{equation}
    \frac{\partial(\varepsilon_s\rho_s)}{\partial t} + \nabla.(\varepsilon_s\rho_s\textbf{u}_s) = D_{eff,s} \nabla^2(\varepsilon_s\rho_s) + \dot{\omega}_s
\end{equation}
\end{subequations}

where,

\begin{subequations}
\begin{equation}
    \varepsilon_g \rho_g = \sum_{k=1}^{N_g} \varepsilon_k \rho_k \ \ \varepsilon_s \rho_s = \sum_{k=1}^{N_s} \varepsilon_k \rho_k
\end{equation}
\begin{equation}
    \varepsilon_g \rho_g \textbf{u}_g = \sum_{k=1}^{N_g} \varepsilon_k \rho_k \textbf{u}_k \ \ \varepsilon_s \rho_s \textbf{u}_s = \sum_{k=1}^{N_s} \varepsilon_k \rho_k \textbf{u}_k
\end{equation}
\begin{equation}
    \dot{\omega}_g = \sum_{k=1}^{N_g} \dot{\omega}_k \ \ \dot{\omega}_s = \sum_{k=1}^{N_s} \dot{\omega}_k
\end{equation}
\begin{equation}
    \dot{\omega}_g + \dot{\omega}_s = \sum_{k=1}^N \dot{\omega}_k = 0 \label{eq:A_w_sum}
\end{equation}
\begin{equation}
    \varepsilon_g + \varepsilon_s = \sum_{k=1}^N \varepsilon_k = 1 \label{eq:A_e_sum}
\end{equation}
\end{subequations}

We only consider the simulation inside perform domain with the following assumptions
\begin{enumerate}
    \item For stationary solid matrix: $\textbf{u}_s = 0 $,
    \item No diffusion by solid species: $D_{k,s} = 0$,
    \item negligible flow and diffusion dominant inside the porous substrate\cite{mcallister1993simulation}: $\textbf{u}_g = 0 $,
    \item gas is in a quasi-steady state, $\frac{\partial(\varepsilon_g\rho_g)}{\partial t} \approx 0$
\end{enumerate}
After applying the above assumptions, the equation simplifies to \\
\begin{subequations}
\begin{equation}
    D_{eff,g} \nabla^2(\varepsilon_g\rho_g) = - \dot{\omega}_g
\end{equation}
\begin{equation}
    \frac{\partial(\varepsilon_s\rho_s)}{\partial t} = \dot{\omega}_s 
\end{equation}
\end{subequations}

Here we redefine $\varepsilon_g\rho_g = MC$, where $C$ is molarity and $M$ is the molar mass, use the equation $\varepsilon_s = 1 - \varepsilon_g$, and assume the reaction to be first-order $\dot{\omega}_s = M KCS_v$. Therefore the governing equations result in the following form.
\begin{equation}
     D_{g,eff} \nabla^2(C) = KCS_v
     \label{eq:Deff}
\end{equation}
\begin{equation}
    \rho_s \frac{d\varepsilon_g}{dt} = - MKCS_v
     \label{eq:eps}
\end{equation}

\section{I-CVI Solver}

\begin{algorithm}[hbt!]
{
\caption{An algorithm for PiNDIff I-CVI solver}\label{alg:ICVI}
  \SetKwFunction{FSum}{PiNDIff-I-CVI}
 
  \SetKwProg{Fn}{Function}{:}{}
  \Fn{\FSum{$C_0(\mathbf{x}), \varepsilon_0(\mathbf{x}), T, P, C_{BC},  \boldsymbol{\theta} = (\boldsymbol{\theta}_{D_{eff}}, \boldsymbol{\theta}_{K}, \boldsymbol{\theta}_{S_v})$}}{
        $t \gets 0$ \\
        \While{$t < t\_max$} 
        { 
            $D_{eff} \gets \mathcal{D}_{nn}\big[\varepsilon_t, T, P; \boldsymbol{\theta}_{D_{eff}}\big]$ \Comment {Compute Diffusion coefficient}\\
            $K \gets \mathcal{K}_{nn}\big[T; \boldsymbol{\theta}_{K}\big]$ \Comment {Compute reaction deposition rate}\\
            $\tilde{S}_v \gets \mathcal{S}_{nn}[\hat{S}_v, \varepsilon_t; \boldsymbol{\theta}_{S_v}]$ \Comment {Compute S2V ratio correction}\\
            $S_v \gets \hat{S}_v (1 + \tilde{S}_v)$ \Comment {Compute S2V ratio}\\
            $C_{t} \gets \text{Elliptic\_Molarity\_Solver}(D_{eff}, K, S_v, C_{BC})$ \Comment {Compute Molarity}\\
            $\varepsilon_{t+1}  \gets \varepsilon_{t} +  \Delta t \frac{M_d}{\rho_d}C_tKS_v$ \Comment {Euler time stepping, (can use RK4)}  
        }
        \KwRet $\big(\{ C(\mathbf{x})\}_{t=0}^{t\_max}, \{ \varepsilon(\mathbf{x}) \}_{t=0}^{t\_max} \big)$\ \Comment {Predicted time series}
  }}
\end{algorithm}

To solve the elliptic PDE Eq.\ref{eq:Deff}, boundary values for molarity are needed. Dirichlet boundary conditions for molarity are applied, and these boundary values are computed based on the partial pressure $P_r$ as 
\begin{equation}
    C_{BC} = \text{mole fraction} \times \text{total moles} = \frac{P_r}{P} \times \frac{P}{RT} = \frac{P_r}{RT}.
\end{equation}
The point-Jacobi method is employed to iteratively solve Eq.\ref{eq:Deff}. While the ODE Eq.~\ref{eq:eps}, needs the initial porosity $\varepsilon_0$ that depends on the preform. The spatiotemporal porosity and molarity $\big(\{ C(\mathbf{x})\}_{t=0}^{t\_max}, \{ \varepsilon(\mathbf{x}) \}_{t=0}^{t\_max} \big)$ is obtained by stepping in time using Euler or RK4 methods. 

\section{Hyper-parameters for PiNDiff I-CVI model}

In the PiNDiff model, the following learning setting is used for the neural network trainable parameters,
\begin{itemize}[noitemsep]
    \item Initial learning rate = $10^{-3}$
    \item Optimizer = Adam
    \item Scheduler = cosine\_decay\_schedule($\alpha=10^{-2}$),
\end{itemize}

\end{document}